%% file: main.tex
\documentclass[journal]{IEEEtran}

\usepackage{xcolor}
\usepackage{graphicx}
\usepackage{amsmath}
\usepackage{amsfonts}
\usepackage{caption}
\usepackage{subcaption}
\usepackage{hyperref}

\usepackage[version=4]{mhchem}
\usepackage{bm}
\usepackage{float}
\usepackage{algorithm}
\usepackage{algcompatible}

\newcommand{\multiline}[1]{%
  \begin{tabularx}{\dimexpr\linewidth-\ALG@thistlm}[t]{@{}X@{}}
    #1
  \end{tabularx}
}

\newcommand{\RETURN}{\text{\bf return}~}

\newcommand{\range}[1]{0, \dots, #1}
\newcommand{\rangeset}[1]{\{\range{#1}\}}

\title{Occlusion-Aware Ground Target Search by a UAV in an Urban Environment}
\author{
\IEEEauthorblockN{
Collin Hague, 
and Artur Wolek}\\
\IEEEauthorblockA{
Department of Mechanical Engineering and Engineering Science\\
Email: \{ chague, awolek\} @charlotte.edu \\
University of North Carolina at Charlotte, \\
Charlotte, NC 28223 USA}}
\begin{document}

\maketitle

\begin{abstract}
This paper considers the problem of searching for a point of interest (POI) moving along an urban road network with an uncrewed aerial vehicle (UAV). The UAV is modeled as a variable-speed Dubins vehicle with a line-of-sight sensor in an urban environment that may occlude the sensor's view of the POI. A search strategy is proposed that exploits a probabilistic visibility volume (VV) to plan its future motion with iterative deepening $A^\ast$. The probabilistic VV is a time-varying three-dimensional representation of the sensing constraints for a particular distribution of the POI's state.  To find the path most likely to view the POI, the planner uses a heuristic to optimistically estimate the probability of viewing the POI over a time horizon. The probabilistic VV is max-pooled to create a variable-timestep planner that reduces the search space and balances long-term and short-term planning. The proposed path planning method is compared to prior work with a Monte-Carlo simulation and is shown to outperform the baseline methods in cluttered environments when the UAV's sensor has a higher false alarm probability.
\end{abstract}

\section{Introduction}

\input{sections/00_Introduction}
\section{Problem Formulation}
\label{sec:problem}

\input{sections/02_ProblemFormulation}
\section{Estimation}
\label{sec:estimation}

\input{sections/03_Estimation}

\section{Path Planning}
\label{sec:planning}
\input{sections/04_PathPlanning}
\section{Numerical Study}
\label{sec:results}

\input{sections/05_Simulation}
\section{Conclusion}
\label{sec:conclusion}
This work proposes a method for searching a space for a moving point of interest (POI) in a dense urban environment with an uncrewed aerial vehicle (UAV). The POI is constrained to a road network, and the proposed method exploits the constraints to create probabilistic visibility volumes (VVs). The probabilistic VVs represent a time-varying three-dimensional scalar field that encodes the likelihood that the UAV can measure the POI with its sensor constrained by line-of-sight and sensing distance.
The UAV plans its trajectory by maximizing the discounted sum of the probability of sensing the POI along the trajectory using iterative deepening $A^\ast$.
A heuristic is proposed to guide the $A^\ast$ search that accounts for the UAV's reachability and probabilistic VVs.
Also, a method of balancing long-term vs short-term planning is created by max-pooling the probabilistic VVs to create a variable-timestep planner by reducing the search space.
The Monte Carlo simulation shows that the proposed method can minimize the search time in dense environments when compared to existing methods for cases where UAVs are equipped with high false alarm rate sensors.
The proposed method leads to a lower search time by adapting its path based on information collected with its sensor, as shown by the lower median search times.
Future work may include implementing the matrix computations on a graphics processing unit (allowing a real-time test flight on a UAV), a statistical test for switching the planner from search to tracking mode, and a method that accounts for the UAV's positional uncertainty when calculating the probability of viewing the POI. Evasive targets or flight at lower altitudes requiring collision avoidance may also be of interest.

\bibliographystyle{elsarticle-num}
\bibliography{main}

\section{Appendix}
\label{sec:appendix}
\input{sections/99_Appendix}
\end{document}

%% file: sections/00_Introduction.tex
Uncrewed aerial vehicles (UAVs) are widely used for airborne sensing tasks that require searching for a target point of interest (POI) with a line-of-sight (LOS) sensor, such as search and rescue, and police and military surveillance \cite{schedl2021autonomous, geiger2008flight}. 
When the UAV searches for a moving POI along a road in an urban environment, the view of the POI may be obstructed by buildings, tunnels, bridges, trees, and other structures.
This work creates a planning and estimation method for a UAV to search for a moving POI.
The road network that constrains the POI's motion is represented as a graph with nodes that are position-velocity pairs, and edges that represent valid transitions between nodes. At each position along the road network, a visibility volume (VV) is constructed, as described in \cite{Hague2023}, to represent the region of airspace in which the UAV can view the target and satisfy both airspace and sensing quality constraints.
The UAV, modeled as a variable-speed Dubins vehicle flying at a constant altitude, carries a camera to measure the POI's position, and uses a recursive Bayesian estimator (RBE) to assimilate noisy position measurements and estimate the POI's state. An advantage of using an RBE, when searching for a moving POI, is that the RBE can accommodate estimation over graph structures and can use ``null'' measurements to update the POI's probability distribution---a null measurement occurs when the UAV does not detect the POI within its sensing region and thus increases the probability of cells not in view.
The path planning method developed in this work uses the estimator's POI state distribution to create a probabilistic VV, which is a time-varying scalar field representing the probability of viewing the POI.
The graph search path planner convolves the probabilistic VV with the set of discretized states attainable after $k$ timesteps, referred to as the UAV's discretized reachability. The resulting convolved quantity is the upper limit of the probability that the UAV will view the POI in $k$ time steps.
The path planner max-pools the probabilistic VVs to reduce the search space, decreasing the computational load and allowing for a larger time horizon.

\subsection{Related Work}
The problem of searching a space for a target involves at least two agents. A \emph{searcher} moves through the space with sensing ability to detect a \emph{target}. The difference between \emph{search} and \emph{tracking} is that a search problem has highly uncertain knowledge of the target's state and must decrease this uncertainty until the target is found, while a tracking problem begins with a relatively accurate estimate of the target's state, and the tracking agent must maintain or improve the estimate over time. The searcher, in this work, is a UAV tasked with finding the target (a moving POI) using a visual LOS sensor. The search problem in our work occurs in an urban environment with visual occlusions and a road network that governs the motion of the POI.

The problem where one or more UAVs search an urban environment with one or more targets has been studied. When in an urban environment, some authors do not account for a sensor's visual occlusion during path planning \cite{Skoglar2012, Meera2019}, while others do \cite{Semsch2009AutonomousEnvironments, Cook2014, Shin2018, Tyagi2021, Jeon2021AutonomousObstacles, Gemerek2022DirectionalAvoidance, Wu2022, Feng2025}. Methods to account for occlusions can be grouped into four categories: (i) casting rays from possible UAV positions to possible target positions \cite{Semsch2009AutonomousEnvironments, Cook2014, Tyagi2021}, (ii) create visibility regions where, for some region of space, LOS constraints for a single point are met \cite{Jeon2021AutonomousObstacles, Gemerek2022DirectionalAvoidance, Hague2023}, (iii) using explicit zones where the target cannot be viewed, e.g., a tunnel \cite{Feng2025}, and (iv) flying directly above possible target locations to avoid
occlusions (assuming no overhead obstacles) \cite{Moore2021, Meng2014}.
This work falls into the second category above and builds upon our previous work with VVs in \cite{Hague2023, Hague2025},
to introduce a novel  \emph{probabilistic VV}, i.e., a scalar field representing the chance of observing the target. The probabilistic VVs enable a path planner to search the environment efficiently.

A UAV searching for a moving target in an urban environment can use knowledge of its dynamics to plan future motion. Fixed-wing UAVs are often modeled as a Dubins vehicle  \cite{Skoglar2012, Cook2014, Feng2025, Dille2014GuaranteedAircraft, Semsch2009AutonomousEnvironments, Sung2016, Ivic2022MotionConditions} or a variable-speed Dubins vehicle \cite{Meng2014, Shin2018}, flying at a constant attitude. Some search strategies ignore such nonholonomic constraints \cite{Wolek.RAL.2020, Jeon2021AutonomousObstacles, Gemerek2022DirectionalAvoidance, Meera2019, Mavrommati2018, Wu2022} while others \cite{Skoglar2012, Cook2014, Feng2025, Dille2014GuaranteedAircraft, Semsch2009AutonomousEnvironments, Sung2016, Ivic2022MotionConditions, Meng2014, Shin2018}, including this work, do not. This work considers the limited turn-rate and non-zero forward velocity of a UAV. 

When planning a path for a UAV to search a space, the motion model of the target influences the choice of planning method. Coverage planners, like the traveling salesperson problem \cite{Hague2023}, and spiral method \cite{Semsch2009AutonomousEnvironments}, whose goal is to efficiently cover the entire space once, are good choices when the target is stationary. However, these methods can fail for a mobile target that transitions from an unsearched space into a previously searched space. Coverage planners, by design, do not purposefully revisit searched space, and are not optimal for searching for moving targets. This leads to a class of problems called \emph{pursuit evasion games} \cite{Dille2014GuaranteedAircraft}. In pursuit evasion games, the target can move with bounded speed or infinitely fast. Solutions to these games often involve a team of searchers systematically exploring the space and guarding previously searched subspaces from the mobile target. The systematic approach guarantees that the target will be found; an example of a search on a road network is \cite{Dille2014GuaranteedAircraft}. In \cite{Dille2014GuaranteedAircraft}, the UAVs always detect the target (without error) and do not need to maintain an estimate of possible target states. Detection failure and false alarms are considered in this work, requiring an estimator to assimilate the target's state, although we do not consider evasive targets.

Many authors loosen the requirement of guaranteeing the search's success and study the mean time to detect the target. 
To minimize the mean detection time, a cost function encoding detection likelihood can be optimized over the UAV's future path. Some works \cite{Hague2023, Gemerek2022DirectionalAvoidance, Semsch2009AutonomousEnvironments} find a set of configurations that meet sensing constraints and then minimize a cost function that considers a path transitioning between the configurations. Other works \cite{Cook2014, Feng2025, Tyagi2021, Jeon2021AutonomousObstacles, Shin2018} encode the sensing constraints into the cost function and search for an optimal sequence of configurations with corresponding controls. Popular optimization techniques for search and tracking are model predictive control/horizon planning  \cite{Skoglar2012, Cook2014, Moore2021, Sung2016, Mavrommati2018, Shin2018, Jeon2021AutonomousObstacles, Tyagi2021} and reinforcement learning \cite{Feng2025, Li2025, Moore2021}. This work uses iterative deepening $A^\ast$ as a horizon planner, similar to Dijkstra’s algorithm in \cite{Moore2021}, by introducing a novel heuristic to guide the search. When the heuristic guides the search, $A^\ast$ does not evaluate as many states as Dijkstra’s algorithm. Moreover, iterative deepening allows a real-time system to interrupt planning before completion, resulting in an optimal path for a smaller time horizon. Iterative deepening can also terminate when a secondary goal is met. 
This work uses a variable resolution grid for planning, similar to \cite{Li2025}, by max-pooling the convolution of the UAV's reachability and the probabilistic VVs. The application of max-pooling decreases the depth of the search tree and reduces the search space, assuming equal time horizons.

A path planner needs an estimate of the target's state distribution to find paths with a high likelihood of viewing the target. Horizon planning methods \cite{Skoglar2012, Cook2014, Moore2021, Jeon2021AutonomousObstacles, Sung2016, Mavrommati2018, Shin2018} propagate the target's state distribution forward in time to calculate the probability of viewing a target at future timesteps. 
One method of calculating state distributions is a grid where grid cells are valid target states with an associated probability \cite{Skoglar2012, Cook2014, Wolek.RAL.2020, Meng2014, Sung2016}. Grids allow for measurement updates when the sensor collects no information, i.e., a ``null'' measurement, in conjunction with the sensing constraints \cite{Skoglar2012, Cook2014, Wolek.RAL.2020, Sung2016}. Grid-based methods also support non-Gaussian probability distributions useful for targets constrained to a road network \cite{Meng2014, Cook2014, Skoglar2012, Wolek.RAL.2020}. The authors in \cite{Moore2021} created a particle filter where the targets were constrained to a road network comprised of line segments as opposed to a grid.
The extended Kalman filter is used for searching in \cite{Mavrommati2018}, but it does not allow the target to be constrained to a road network.
The estimator in this work is formulated as a grid-based recursive Bayesian estimator (RBE) that is similar to other grid-based planners in \cite{Skoglar2012, Cook2014, Wolek.RAL.2020, Meng2014, Sung2016}, but it considers a 3D representation of visibility to calculate the probability of viewing a target at a particular grid cell. The target motion model in this work is an implementation of the ``decelerate before turning'' model, in the tracking problem \cite{Shin2018}, as a discrete Markov chain.

\subsection{Contributions}
The contributions of this work are: (i) the concept of probabilistic VVs and an algorithm to efficiently compute and  represent them as matrices, (ii) a heuristic to approximate the upper bound on the probability of observing the uncertain POI for some future time horizon that considers the UAV's reachability, and (iii) a method for balancing short-term and long-term planning by applying max-pooling to the probabilistic VVs.
A Monte-Carlo simulation compares the proposed method to an existing VV-based static point coverage planner adapted for the search problem and a non-VV-based lawnmower coverage planner.

\subsection{Paper Organization}
The remainder of this paper is organized as follows. Section~\ref{sec:problem} introduces the environment model, vehicle motion model, POI motion model, measurement model, and the problem statement. Section~\ref{sec:estimation} describes the estimator that predicts the distribution of future POI states. Section~\ref{sec:planning} creates a path planning strategy for the UAV to find the POI. Section~\ref{sec:results} presents several illustrative examples and results from a Monte-Carlo study. Section~\ref{sec:conclusion} concludes the paper.

%% file: sections/02_ProblemFormulation.tex
This section formulates the problem of estimating the state of a moving target through a dense urban environment. 
The UAV's motion is approximated by a variable-speed Dubins vehicle at a constant altitude, and its environment consists of 3D obstacles and a road network. The section introduces the POI's motion on the road network, the UAV's sensor model, and the UAV's measurement of the POI. Lastly, the problem statement is formulated to maximize the cumulative probability of viewing the POI over the mission.

\subsection{UAV Motion Model}
Consider a fixed-wing UAV with planar position $(q_x, q_y)$, heading $q_\psi$, and constant altitude $z=h_{\rm UAV}$ modeled as a variable-speed Dubins vehicle,
\begin{equation}
    \begin{bmatrix}
        \dot q_x\\\dot q_y \\\dot q_\psi
    \end{bmatrix}= \begin{bmatrix}
        u_v\cos\psi\\u_v\sin\psi\\u_\psi
    \end{bmatrix}\;,
    \label{eq:dynamics}
\end{equation}
where $\bm{q}=[q_x, q_y,q_\psi]^{\rm T}\in {\rm SE}(2)$ is the system's state, $u_v$ is the velocity control, and $u_\psi$ is the steering control. 
The velocity control is bound by 
\begin{equation}
    \label{eq:speed_constraint}
    u_{v, {\rm min}} \leq u_v \leq u_{v, {\rm max}}\;,
\end{equation}
where the bounding speeds are $0 < u_{v, {\rm min}} \leq u_{v, {\rm max}}$.
The steering control,
\begin{equation}
\label{eq:turn_constraint}
    |u_{\psi}| \leq u_{\psi,\rm max}\;,
\end{equation} 
is bounded by the maximum turn-rate $u_{\psi, \rm max}$.
\subsection{Environment Model}
\label{sec:environment}
\begin{figure}[t]
    \centering
    \includegraphics[height=1.3in]{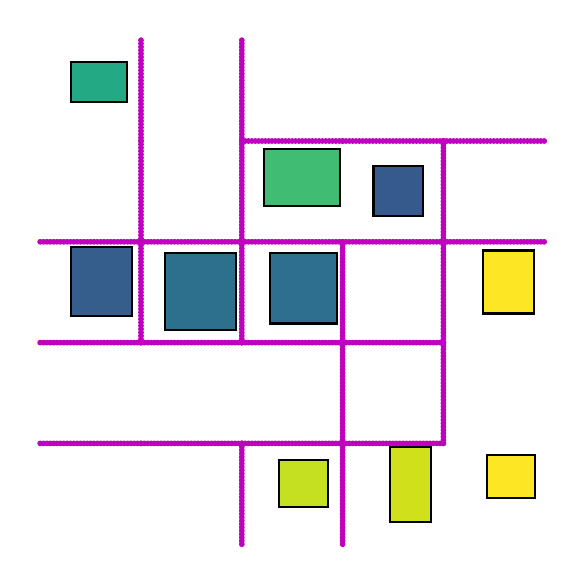} ~
    \includegraphics[height=1.3in]{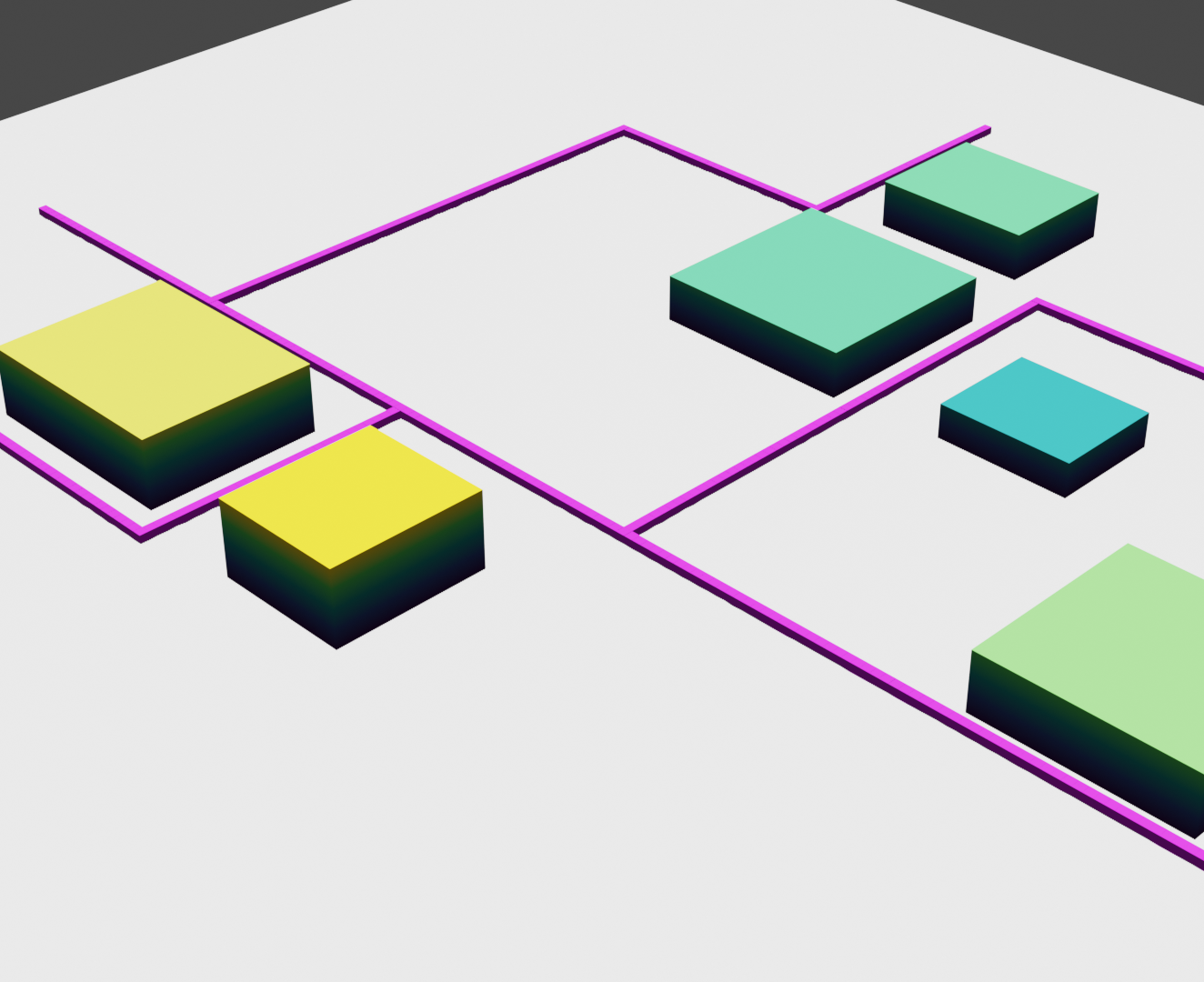}
    \caption{Example of the environment where a UAV tracks a point of interest, left: 2D view and right: 3D view.}
    \label{fig:environment}
\end{figure}

The UAV operates in an urban environment that consists of a ground plane with a road network, $G_0$, and a collection of three-dimensional objects representing buildings or other structures, $B$.
The $i$th object, in the set $B$ of $N_O $ objects, is an extruded polygon $B_i = \{(x,y,z) \in \mathbb{R}^3~|~(x,y) \in A_i$ and $z \in [0, h_i]\}$, where $A_i$ is a simple two-dimensional polygon, and $h_i$ is the height of the obstacle.
The boundary of polygon $A_i$ is denoted $\partial A_i$, whose shape is defined by an ordered set of points with a positive signed area.
The set of points on the interior of $A_i$ is denoted ${\rm int}(A_i)$. 
The polygonal areas of each object do not intersect ${\rm int}(A_i) \cap {\rm int}(A_j) = \emptyset$ for all $i\neq j$ with $i,j \in \rangeset{N_O-1}$.
The polygons rest on the ground plane, $z=0$ (elevation changes are not considered). An example environment is shown in Fig.~\ref{fig:environment}, where the buildings are colored purple to yellow based on increasing height, the road network is magenta, and the ground plane is white.

The road network in the urban environment is defined as an undirected graph $G_0$ with a set of nodes $\mathcal{N}_0 \in \mathbb{Z}$ and bidirectional edges $\mathcal{E}_0 \in \mathcal{N}_0 \times \mathcal{N}_0$. Each node is assigned a position along a road $\bm{g}\in\mathcal{G}_0$ on the ground plane, where $\mathcal{G}_0$ is the set of positions corresponding to $\mathcal{N}_0$. The edges $(n_i, n_j) \in \mathcal{E}_0$ are defined between adjacent nodes $n_i, n_j \in \mathcal{N}_0$ when a road connects them. The moving POI travels along the edges between nodes on the graph, which approximates the POI traveling along a road. The edges cannot intersect obstacles, $d(\tau;\bm{g}_i, \bm{g}_j) \cap {\rm int}(A_k) = \emptyset$ for all $(n_i, n_j) \in \mathcal{E}_0$ and $k \in \rangeset{N_O - 1}$ where $d(\tau;\bm{g}_i, \bm{g}_j) = \{\bm{g}_i + \tau(\bm{g}_j - \bm{g}_i)~{\rm for}~\tau \in [0,1]\}$ is the line segment between the points. 
To construct the graph $G_0=(\mathcal{N}_0,\mathcal{E}_0)$, start with the road network in Fig.~\ref{fig:environment}. At each intersection or dead end, place a node. For each road between two nodes, place an edge. 

\begin{figure}[t]
    \centering
    \includegraphics[width=.23\textwidth]{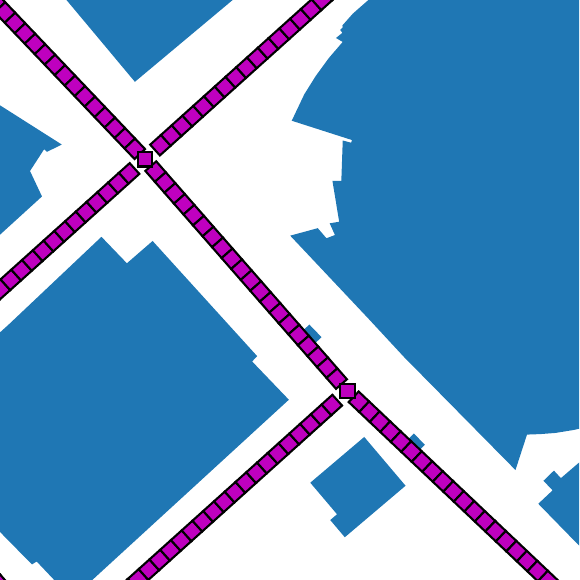}~
    \includegraphics[width=.23\textwidth]{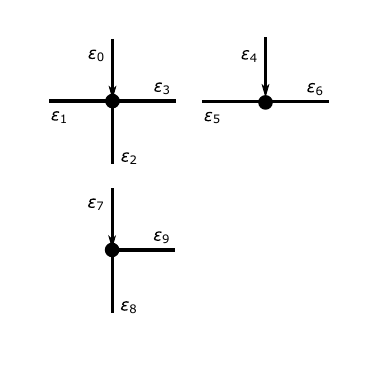}
    \caption{A road network in an urban environment and the types of intersection in the environment.}
    \label{fig:cells}
\end{figure}

While the graph $G_0$ represents the geometry of the road network, a higher resolution \emph{target estimation graph} $G_1 =(\mathcal{N}_1, \mathcal{E}_1)$ is used to estimate the target's position. To construct, $G_1$, first create the corresponding set of points $\mathcal{G}_1 \subset \mathbb{R}^2$ as the union of a the road network points $\mathcal{G}_0$ and a set of points that are sampled along the edges $\mathcal{E}_0$ of the graph $G_0$ with a separation distance $l_c$. As the road network edges are discretized, the adjacency of neighboring nodes is recorded. To allow for estimating both position and velocity, the target estimation graph $G_1$ is a directed graph with a nodes $\mathcal{N}_1 \subset \mathbb{R}^4$ that correspond to position-velocity pairs and edges $\mathcal{E}_1 \subset \mathcal{N}_1 \times \mathcal{N}_1$ that join nearby points with different velocities. A set of $m$ positive discretized speeds $V =\{v_0, \ldots v_{m - 1}\}$ is considered, creating discretized velocities where the direction is defined by the relative geometry of adjacent points. The nodes  $\mathcal{N}_1$ are constructed by considering each point in $\mathcal{G}_1$ and whether it is at an intersection (i.e., in $\mathcal{G}_0$) or not. For each point not on an intersection, $\bm{g} \notin \mathcal{G}_0$,  let ${\bm g}_i \in \mathcal{G}_0$ and ${\bm g}_j \in \mathcal{G}_0$ denote the two points on the road network edge whose line segment was sampled to produce the node at ${\bm g}$. From these points the $m$ velocities $v_a(\bm{g}_i - \bm{g}_j) /||\bm{g}_i - \bm{g}_j||$ are defined, along with their negative counterparts, where $v \in V$, to give a total of $2m$ velocities corresponding to ${\bm g}$. Joining these velocities with ${\bm g}$ gives $2m$ nodes to be added to $\mathcal{N}_1$. Now, for points ${\bm g} \in \mathcal{G}_1$ that are at intersections (i.e., ${\bm g} \in \mathcal{G}_0$) a different procedure is used. Since ${\bm g}$ is at an intersection, it has a corresponding node $n \in \mathcal{N}_0$ in the road network. Let $\mathcal{I}(n) = \{ n' \in\mathcal{N}_0~{\rm such~that}~(n,n') \in \mathcal{E}_0\}$ be the set of neighbors of $n$ in the road network, and let $|\mathcal{I}(n)|$ be the cardinality of this set. Two velocities are considered for each neighbor (both moving away from and towards the intersection) in a manner defined similar as before. Thus, the total number of nodes to be added for each intersection point ${\bm g}$ is $2m|\mathcal{I}(n)|$. Edges are defined between each part of nodes in $\mathcal{N}_1$ who have positions that are adjacent in the sense of their proximity in how the point set $\mathcal{G}_1$ was constructed. $G_1$ contains self-loops when a speed of zero is considered.

An example environment with the graph $G_1$ is shown in Fig.~\ref{fig:cells}, where the blue polygons are the building footprints and the magenta squares are the road network.
To avoid collisions, the UAV's flight path is constrained to a feasible airspace above obstacles. The feasible airspace  $F = \mathbb{R}^2 \times (h_{\rm building}, h_{\rm feasible})$ for the UAV is the region of space between the height of the tallest building, $h_{\rm building}$, and the maximum feasible altitude, $h_{\rm feasible}$.

\subsection{Point of Interest Model}
\label{sec:POI_model}
The state of the POI is constrained to the graph $G_1$ and moves along the edge from one node to the next according to a discrete Markov chain \cite{bell2013}.  The path traversed by the POI, $J = \{ (n_k, t_k)~{\rm such~that}~k \in \rangeset{N_{k} - 1}~{\rm and}~n_k \in \mathcal{N}_1\}$, is denoted by a sequence of nodes $n_k$ with corresponding times $t_k$, and the number of timesteps $N_k$. For a path to be valid, the sequence of nodes on the graph $G_1$ must have a valid sub-path of $v_k / l_c$ nodes between the nodes $n_k$ to $n_{k + 1}$. The difference $t_{k + 1} - t_k= \Delta t$ is a constant value. The discrete Markov chain \cite{bell2013} that models the POI's motion as a car driving through a city is:
\begin{equation}
    \label{eq:markov}
    \zeta: s_k \rightarrow s_{k + 1}\;,
\end{equation}
which maps the POI's current state $s_k \in \mathcal{N}_1$ to the next state $s_{k + 1} \in \mathcal{N}_1$. The planar location of the states $s_k$ is denoted $\bm{g}_{s, k} \in \mathcal{G}_1$ and the planar velocity $\dot{\bm{g}}_{s, k}$. The discrete Markov chain $\zeta$ has the property that the probability of transitioning to the next node $s_{k + 1}$ only depends on the previous state $s_k$ and is denoted $\pi(s_k, s_{k + 1})$. Since the POI's motion model only depends on the previous state, the state transition probabilities can be written as a matrix $\bm{Z} \in [0, 1]^{\delta \times \delta}$ where $\delta=|\mathcal{N}_1|$ is the cardinality of the set $\mathcal{N}_1$. The elements of the row stochastic matrix ${\bm Z}$ are $z_{ij}=\pi(s_i, s_j)$, the probability of transitioning from state $s_i$ to state $s_j$.

When the car reaches an intersection, the probability of turning depends on the current speed and the intersections shown in Fig.~\ref{fig:cells}. There are three types of intersections in Fig.~\ref{fig:cells}, four-way (top left), three-way fork (top right), and three-way with a straight direction of travel (bottom left).
When there is no intersection, the POI moves forward. If there is a dead end, it turns around.
When the POI is approaching or leaving an intersection, the vehicle has an increased probability of decelerating or accelerating, respectively. The POI also has a chance of remaining at the same speed. The POI only changes velocity in steps of $l_c/\Delta t$. 
When a target approaches an intersection, the probability of decelerating increases with decreasing distance to the intersection. If the POI cannot decelerate to the slowest speed at the intersection, then the POI is more likely to accelerate. Once the POI passes the intersection, the probability of decelerating decreases, and the POI is more likely to accelerate. All of these behaviors are encoded by appropriately defining $\pi(s_i, s_j)$.

\subsection{Measurement Model}
\label{sec:measurement}
The UAV is equipped with an imaging sensor (e.g., RGB, infrared, or hyperspectral camera) with a 360$^\circ$ field of view that is used to detect the POI.
The sensor data is processed in real-time to produce an inertial position measurement $\bm{\xi} \in \mathbb{R}^2$ of the POI. The measured position is not constrained to be precisely on the road network.

A visibility region is a region of space where a point on the ground plane $\bm{g}\in \mathbb{R}^{2}$ can be observed from any point $\bm{\rho} \in F \subseteq \mathbb{R}^3$ in the feasible airspace. 
The line segment $d(\tau; \bm{g}, \bm{\rho})$
is used to define a \emph{visibility volume} (VV), which is a set of points that have a direct LOS to the target (i.e., not obscured by buildings). The VV is also constrained by a maximum range $l_{\rm max}$, the minimum imaging requirement in number of pixels for a target of known size with known camera resolution. The VV for a target located at ${\bm g}$ is the subset of the feasible airspace $F$ that is within direct LOS to the target, and within a maximum range $l_{\rm max}$ relative to the target,
\begin{equation}
    \begin{split}
        \alpha({\bm g};F, l_{\rm max}, B) = &
        \{ {\bm \rho} \in F ~\text{for}~
        ||{\bm \rho} - [g_x, g_y, 0]^{\rm T} || \leq l_{\rm max} \\ &\text{and}~
         d(\tau; {\bm g}, {\bm \rho}) \cap B_j = \emptyset~\text{for all}~\tau \in[0,1]\\ &\text{and}~j \in \rangeset{N_O-1} \}\;.
        \end{split}
    \label{eq:visibility_volumes}
\end{equation}
For brevity, visibility volumes \eqref{eq:visibility_volumes} are henceforth denoted $\alpha({\bm g})$.
VVs are computed using \cite{Hague2023}, resulting in a triangular mesh.

The UAV's sensor is imperfect and may fail to \emph{detect} the POI when the UAV is inside the POI's VV with constant probability $p_d$. If the imaging sensor detects the POI, the planar position $\bm{g}_s$ is measured with additive uncorrelated two-dimensional zero-mean Gaussian noise $\bm{v}\sim\mathcal{N}(\bm{0}, \bm{R})$ where $\bm{R} \in \mathbb{R}^{2\times 2}$ is sensor noise covariance matrix. If the UAV is not in the POI's visibility volume, or the sensor fails to detect the POI, then the imaging sensor creates a \emph{null} measurement, $\emptyset$. The sensor may also erroneously create a \emph{false alarm} measurement that is a random valid POI position $\bm{g}_f \in \mathcal{G}_1$ inside the UAV's sensing region $\bm{q}\in\alpha(\bm{g}_{f})$ with constant probability $\mu$.
The measurement equation for sensing the POI, considering the VVs is
\begin{equation}
    \label{eq:sensor}
    h(s^\ast, \bm{q}; r_d, r_f, p_d, \mu) = \begin{cases}
        \bm{g}_f + \bm{v} & {\rm if}~ r_f \leq \mu\\
        \bm{g}_{s^\ast} + \bm{v} & {\rm else~if}~\bm{q} \in \alpha(\bm{g}_{s^\ast})~{\rm and}~r_d < p_d\\
        \emptyset & {\rm else~if}~ \bm{q} \in \alpha(g_{s^\ast})~{\rm and}~r_d \geq p_d\\
        \emptyset & {\rm otherwise}
    \end{cases}\;,
\end{equation}
where $s^\ast \in \mathcal{G}_1$ is the true state of the POI, and $r_d\sim U(0, 1)$ and $r_f\sim U(0, 1)$ are two uncorrelated random variables from the random uniform distribution between $0$ and $1$ inclusive. The first case (false positive) $r_f \leq \mu$ is when the sensor creates an erroneous measurement $\bm{g}_f + \bm{v}$ of a random possible POI state $\bm{g}_f \in \mathcal{G}_1$ visible from the UAV $\bm{q} \in \alpha(\bm{g}_f)$. The second case (true positive) occurs when $\bm{q} \in \alpha(\bm{g}_{s^\ast})~{\rm and}~r_d < p_d$ allowing the UAV to measure the POI's location $\bm{g}_{s^\ast} + \bm{v}$ because the UAV is inside the POI's VV and the UAV detects the POI. The third case (false negative) $\bm{q}\in \alpha(\bm{g}_{s^\ast})$ occurs when the UAV is in the POI's VV and the UAV does not detect the POI $r_d \geq p_d$ creating a null measurement, $\bm{\xi} = \emptyset$. The fourth case (true negative) occurs when the UAV is not in the POI's visibility volume $\bm{q}\notin\alpha(\bm{g}_{s^\ast})$ and the sensor does not create a false alarm measurement $r_d > \mu$ resulting in a null measurement, $\bm{\xi} = \emptyset$.
The imaging sensor operates once every timestep $t_k$ creating a measurement $\bm{\xi}_k=h(s_k^\ast, \bm{q}_k; r_{k, d}, r_{k, f}, p_d, \mu)$, and the noisy measurement history of the sensor is denoted $\bm{\xi}_{0:k}=\{\bm{\xi}_0,\ldots,\bm{\xi}_k\}$ where $k$ is the current timestep.

\subsection{Problem Statement}
The goal of this work is to find a series of turn-rate and speed control inputs $\{\bm{u}_0(t), \ldots \bm{u}_{N_k - 1}(t)\}$ for the variable-speed Dubins vehicle that satisfies constraints \eqref{eq:speed_constraint}--\eqref{eq:turn_constraint} and maximizes the probability of detecting the POI over the time horizon $T = \{0,\ldots,t_{N_k - 1}\}$
where $N_k$ is the number of timesteps. 
The probability of detecting the POI at UAV configuration $\bm{q}_k$ at time $t_k$ is
\begin{equation}
    p_{\rm view}(\bm{q}_k) = \sum_{i=0}^{\delta - 1} (1 - \mu)p_df(\bm{q}_k, \bm{g}_{s,i})p(s_i| \bm{\xi}_{0:k-1})
    \label{eq:prob_detection}\;,
\end{equation}
where $\bm{q}_k$ is the configuration of the UAV at time $t_k$, $1-\mu$ is the probability of no false alarms, $p_d$ is the probability of detection, $p(s_i|\bm{\xi}_{0:k-1})$ is the probability density function of a possible POI location $s_i$ given a measurement history $\bm{\xi}_{0:k-1}=\{h(s^\ast_0, \bm{q}_0),\ldots,h(s_{k-1}^\ast,\bm{q}_{k-1})\}$, and the function
\begin{equation}
    f(\bm{q}, \bm{g}_s) = \begin{cases}
        1 & {\rm if}~[q_x, q_y, h_{\rm UAV}]^{\rm T} \in \alpha(\bm{g}_s) \\
        0 & {\rm otherwise}
    \end{cases}\label{eq:visibile}
\end{equation}
returns $1$ if the UAV is in the VV at the position $\bm{g}_s$ for state $s$ and $0$ otherwise.
While \eqref{eq:prob_detection} is the probability of detecting the target at one time instant, the optimization considers detection over a time horizon starting at the current time step $k$ over $N_k-1$ time steps into the future by maximizing the cumulative discounted probability:
\begin{equation}
    \label{eq:problem}
    {\rm maximize}\quad  \sum_{j \in T}\gamma^j p_{\rm view}(\bm{q}_{k+j})\;,
\end{equation}
subject to the UAV's motion model \eqref{eq:dynamics}--\eqref{eq:turn_constraint}, the POI's motion model \eqref{eq:markov}, the UAV's sensing model \eqref{eq:sensor}, and the initial conditions of the UAV $\bm{q}_0=[q_x, q_y, q_\psi]^{\rm T}$, where $\gamma < 1$ is a discounting factor that weight rewards in the shorter term more heavily than rewards further into the future. The maximization \eqref{eq:problem} is over a receding horizon. The probability of viewing the target $p_{\rm view}(\bm{q}_k)$ changes every timestep when a new measurement $\bm{\xi}_k$ is taken, requiring the planner to re-plan every timestep.

%% file: sections/03_Estimation.tex
The UAV estimates the POI's state by collecting noisy measurements with a LOS sensor \eqref{eq:sensor}, and incorporating knowledge of the road network $G_1$ and motion model \eqref{eq:markov} through a recursive Bayesian estimator \cite{bell2013}. The resulting POI's state distribution on the road network  $p(s|\bm{\xi}_{0:k})$ at time $t_k$ is then used in \eqref{eq:prob_detection} and \eqref{eq:visibile} for optimization.

\subsection{Probability Distributions as Matrices}
\label{sec:vvmatrix}
\begin{figure}
    \centering
    \includegraphics[width=3.25in]{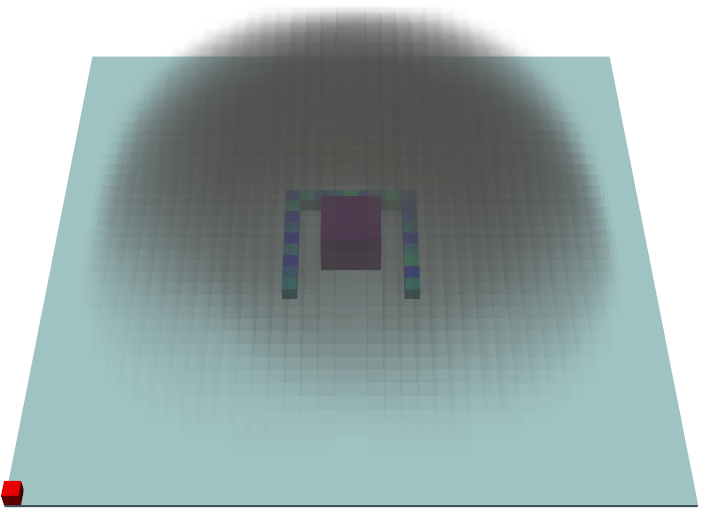}
    \caption{A probabilistic visibility volume for a ``U'' shaped road network around a single building.}
    \label{fig:probVV}
\end{figure}

Previously, the work \cite{Hague2023} approximated VVs as triangular meshes; Here, the VVs are approximated by a tensor $\bm{\alpha}^\square(\bm{g})$ where $\alpha^\square_{ijk}(\bm{g}) \in \{0, 1\}$ is a binary variable and $i,j,k$ are the indices of a set of 3D voxels where each voxel's center is inside the VV. The sets of voxels are created in a common coordinate system, such that each voxel corresponds to an index $i, j, k \in \{0, \ldots, n-1\}^3$ where $n^3$ is the number of voxels in the environment. The side length of a voxel is $l_v$. The left-front-bottom most corner of the environment is the origin with coordinate $[O_x, O_y, O_z]^{\rm T}$ in inertial space and index $[0, 0, 0]^{\rm T}$, the red cube in Fig.~\ref{fig:probVV}.  The value of $\alpha^\square_{ijk}(\bm{g})$ is one if the voxel is entirely inside the VV and zero if the voxel has at least some part outside the VV. The intersection of visibility volumes $\alpha(\bm{g}_0) \cap \alpha(\bm{g}_1)$ is  $\bm{\alpha}^\square(\bm{g}_0) \circ \bm{\alpha}^\square(\bm{g}_1)$ where $(\cdot)\circ(\cdot)$ is the Hadamard or element-wise product. The union of visibility volumes $\alpha(\bm{g}_0) \cup \alpha(\bm{g}_1)$ is ${\rm sat}(\bm{\alpha}^\square(\bm{g}_0) + \bm{\alpha}^\square(\bm{g}_1), 0, 1)$ where ${\rm sat}(\bm{\alpha}^\square, a, b)$ is the saturation of the elements of the tensor $\bm{\alpha}^\square$ to a minimum value of $a$ and a maximum value of $b$.
Using the set of all possible POI states $\mathcal{N}_1$, the probability of a particular state given a measurement history, $p(s|\bm{\xi}_{0:\tau})$, and the corresponding visibility volumes $\bm{\alpha}^\square(\bm{g}_{s, \tau})$ where, to avoid confusion with the voxel index $k$, we temporarily use $\tau$ to represent the current timestep, the probabilistic VV or the scalar field representing the probability of viewing the POI is,
\begin{equation}
    \bm{A}^\square = \sum_{i=0}^{\delta -1}p(s_i|\bm{\xi}_{0:\tau})\bm{\alpha}^\square(\bm{g}_{s, i})\;,
\end{equation}
where each element of the tensor $\bm{A}^\square_{ijk}$ is a Bernoulli random variable. The probability of viewing the POI at the center of the associated voxel is shown in Fig.~\ref{fig:probVV} as the transparent to opaque black cubes. The color of the road network, blue to green, signifies the, low to high, probability that the POI is at that road network node. The purple rectangle is an occluding obstacle.

\begin{figure}
    \centering
    \includegraphics{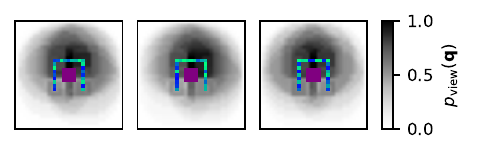}
    \caption{A matrix representing the probabilistic visibility for a ``U'' shaped road network.}
    \label{fig:probMat}
\end{figure}

Since the system \eqref{eq:dynamics} remains at a constant altitude $h_{\rm UAV}$ there is a matrix $\bm{\mathfrak{a}} \in \{0, 1\}^{n\times n}$ that approximates a VV at the altitude $h_{\rm UAV}$, with elements $\bm{\mathfrak{a}}_{ij}=\bm{\alpha}^\square_{ijk}$ where the $z$ index $k=k_h$ respects the inequality
\begin{equation}
    \label{eq:voxelHeight}
    k_hl_v\leq h_{\rm UAV}\leq(k_h + 1)l_v\;.
\end{equation}
There also exists a matrix $\bm{\mathfrak{A}} \in [0, 1]^{n\times n}$ that approximates the probabilistic visibility volume at the altitude $h_{\rm UAV}$, with elements $\bm{\mathfrak{A}}_{ij}=\bm{A}^\square_{ijk}$ and the $z$ index $k=h_h$ also respected \eqref{eq:voxelHeight}.
Figure~\ref{fig:probMat} shows example matrices approximating a probabilistic VVs at height $h_{\rm UAV}$, where the color white to black shows an increasing chance of viewing the POI.

\subsection{Recursive Bayesian Estimator}
An RBE estimates the state of the target given measurements \eqref{eq:sensor}. The RBE is defined as,
\begin{equation}
    p(s| \bm{\xi}_{0:k}) = \frac{p(\bm{\xi}_k|s)}{p(\bm{\xi}_k|\bm{\xi}_{0:k - 1})}p(s|\bm{\xi}_{0:k- 1})\;, \label{eq:bayes_lay}
\end{equation}
where $s \in \mathcal{N}_1$ is the state space of the estimator, the posterior estimate is $p(s|\bm{\xi}_{0:k})$ (the probability distribution of the location of the POI given the measurement history), the measurement likelihood is $p(\bm{\xi}_k|s)$ (the probability of the measurement given a possible POI state), the motion prediction is $p(s|\bm{\xi}_{0:k-1})$ (the probability of the current position based on the previous measurements), and the normalizing factor is $p(\bm{\xi}_k|\bm{\xi}_{0:k-1})$ (the probability of the current measurement given all of the previous measurements).

For a sensor that always detects the POI (when the sensing constraints are met) and does not have false alarms, the likelihood of measurement $\bm{\xi}$ given state $s$ is
\begin{equation}
    L(\bm{\xi} | s) = \begin{cases}
        \eta(\bm{\xi}|s) & {\rm if}~\bm{\xi}\neq\emptyset\\
        1 - f(\bm{q}, \bm{g}_s) & {\rm otherwise}
    \end{cases}\;,
    \label{eq:likelihood_ideal}
\end{equation}
where $f(\bm{q}, \bm{g}_s)$ is \eqref{eq:visibile} and describes the visibility of the POI state $s$ at position $\bm{g}_s$ from $\bm{q}$, and
\begin{equation}
    \eta(\bm{\xi}| s)=
        \frac{e^{\frac{1}{2}(\bm{\xi} - \bm{g}_s)^{\rm T}\bm{R}^{-1}(\bm{\xi} - \bm{g}_s)}}{ \sqrt{(2\pi)^2\det{\bm{R}}}}l_c^2\;,
        \label{eq:gaussian}
\end{equation}
is the probability of a measurement $\bm{\xi}$ originating from a square region, with side length $l_c$ around a possible POI state $s$ according to a multivariate Gaussian distribution. Equation \eqref{eq:gaussian} is a multivariate Gaussian distribution integrated over the square region surrounding the possible POI state $s$ using a midpoint approximation.

\sloppy This work adopts a modified form of a  likelihood function described in \cite{bell2013} where the the authors introduced $L(\bm{\Xi}|s) = p(\bm{\Xi}|s)$, that accounts for a sensor that creates $m$ different measurements $\bm{\Xi} = \{\bm{\xi}^0,\ldots,\bm{\xi}^{m - 1}\}$ at a single instance of time $t_k$, the probability of detection is $p_f(s)$, and the binomially distributed probability of false alarm is
\begin{equation}
    {\rm Pr}(i;m) = \frac{m!}{i!(m-i)!}\mu^i(1 - \mu)^{m -i}\;,
    \label{eq:poisson}
\end{equation}
where $i$ is the number of false alarms at a single point in time, $m$ is the maximum number of false alarms possible in a single timestep, and $\mu$ is the constant probability of a single sensor measuring a false alarm. Since this work considers a sensor that measures the POI only once per timestep then $m=1$; However, we first explain the approach in \cite{bell2013} for $m \geq 1$ before specializing it to our case.
The probability distribution of the $i$th measurement, from the true positive measurement distribution is $\eta(\bm{\xi}_i | s)$, and from the false alarm distribution is $w(\bm{\Xi})$. The likelihood function is \cite{bell2013},
\begin{equation}
\begin{split}
    L(\bm{\Xi} | s) & = \frac{1}{m}\sum_{i=0}^{m - 1}\left[p_f(s)\eta(\bm{\xi}_i|s){\rm Pr}(m - 1; m)w(\bar{\bm{\Xi}}_{{i}})\right] \\ & + \left(1 - p_f(s)\right){\rm Pr}(m;m)w(\bm{\Xi})\;.
    \end{split}
    \label{eq:likelihood_multisensor}
\end{equation}
The term $p_f(s) \eta(\bm{\xi}_i|s) {\rm Pr}(m - 1;m) w(\bar{\bm{\Xi}}_i)$ in \eqref{eq:likelihood_multisensor} is the probability that only the $i$th measurement is a true measurement of the state $s$ and all of the other measurements are false alarms. The notation $\bar{\bm{\Xi}}_i$ denotes the set $\bm{\Xi} \backslash \bm{\xi}_i$, the set of all measurements without the $i$th measurement. The term $\left(1 - p_f(s)\right){\rm Pr}(m;m)w(\bm{\Xi})$ in \eqref{eq:likelihood_multisensor} is the probability that the target was not detected and all of the measurements are false alarms.

For the sensor \eqref{eq:sensor}, the probability of detection is $p_f(s) = p_df(\bm{q}, \bm{g}_s)$ where $p_d$ is the fixed probability of detection and $f(\bm{q}, \bm{g}_s)$, described by \eqref{eq:visibile}, determines whether the system \eqref{eq:dynamics} is inside or outside of the VV at $\bm{g}_s$ for POI state $s$. When the sensor creates a false alarm measurement, all states visible from the configuration $\bm{q}$ are equally likely to have created the measurement, thus the probability that a given state $s$ generates a false alarm measurement is
\begin{equation}
    \Gamma(s) = \frac{f(\bm{q}, \bm{g}_s)}{\sum_{i=0}^\delta f(\bm{q}, \bm{g}_{s, i})}\;,
    \label{eq:fapdf}
\end{equation}
since only visible states can produce a measurement.
From \eqref{eq:sensor}, a false alarm measurement is the planar position of a random visible POI state $\bm{g}_s$ with additive Gaussian noise; Therefore, the probability that the visible POI state $s$ created the false alarm is $\eta(\bm{\xi} | s)\Gamma(s)$. The probability that $\bm{\xi}$ was realized from the false alarm distribution when $\bm{\xi}\neq \emptyset$ and $m=1$ is
\begin{equation}
    w(\bm{\xi}) = \sum_{i=0}^{\delta -1}\eta(\bm{\xi} | s_i)\Gamma(s_i)\;.
    \label{eq:falsealarm_dist}
\end{equation}
By substituting $m=1$, $p_f(s)=p_df(\bm{q},s)$, ${\rm Pr}(m-1;m)=1 -\mu$, ${\rm Pr}(m;m)=\mu$, and \eqref{eq:falsealarm_dist} into \eqref{eq:likelihood_multisensor}, the likelihood function that approximates $L(\bm{\xi}_k, s_k) \approx p(\bm{\Xi}_k|s_k)$ when $\bm{\xi} \neq \emptyset$ is,
\begin{equation}
\begin{split}
    L(\bm{\xi}| s) & = p_d f(\bm{q}, \bm{g}_s)\eta(\bm{\xi} | s)(1 - \mu)  + \mu w(\bm{\xi})\;.
    \label{eq:likelihood}
\end{split}
\end{equation}
Comparing \eqref{eq:likelihood_multisensor} to \eqref{eq:likelihood}, the first term does not include $w(\bar{\bm{\Xi}}_{i})$ because the vector $\bar{\bm{\Xi}}_{i}$ is the set of all measurements without the $i$th measurement and is empty when $m = 1$. The second term does not include the probability of non-detection because the sensor \eqref{eq:sensor} has an error probability independent of detecting the POI.
When the sensor does not detect the POI, $\bm{\xi}=\emptyset$, the likelihood function is,
\begin{equation}
    L(\emptyset | s) = (1 - \mu)(1 - p_df(\bm{q}, \bm{g}_s))\;,
\end{equation}
where false alarm measurements have zero probability because false alarms always create a non-null measurement.

%% file: sections/04_PathPlanning.tex
To minimize $\eqref{eq:problem}$, the future path of the UAV and POI must be considered.
To find the path that maximizes the probability of detecting the POI, the UAV must consider where it can reach and the likelihood of viewing the POI at reachable locations.

\subsection{Discretized Reachability}
Reachability in control theory refers to the set of states that can be reached with feasible control inputs over a time interval $t\in [0, t_f]$ when starting from an initial set at time $t=0$. In this work, the \emph{discretized reachability} of a system has a related but different meaning; It is the set of voxels that can be reached precisely at the time $t = t_f$ starting from an initial state $\bm{q}_0 = [0, 0, \psi_0]^{\rm T}$ at time $t_0$. The discretized reachability of the system \eqref{eq:dynamics} can be described as a set of voxels and, in turn, a matrix. Based on the maneuverability of system \eqref{eq:dynamics}, choose the length of the voxels $l_v$ (the same length as the VV voxels in Sec.~\ref{sec:vvmatrix}, $l_c=l_v$) and the discretized set of headings $\Psi =\{0,\Delta\psi,\ldots, 2\pi - 2\Delta\psi,2\pi - \Delta\psi\}$ where $N_\psi\Delta \psi = 2\pi$, and $N_\psi = |\Psi|$.
The discretized reachability is found by solving for Dubins paths from an initial condition of $\bm{q}_0=[0, 0, \psi_0]^{\rm T}$, where heading $\psi_0 \in \Psi$, to the center point of any voxel with final heading $\psi_1 \in \Psi$. 
The matrix describing the discretized reachability set in the $xy$ plane from initial state ${\bm q}_0$ and for $k$ discrete actions, each of duration $\Delta t$, is denoted $\bm{R}_k(\psi_0)\in \{0, 1\}^{(2w + 1) \times (2w + 1)}$ where $w=\lceil u_{v, \rm max} k\Delta t/l_v\rceil$ and $\bm{q}_0$ is in the center of $\bm{R}_k(\psi_0)$ with corresponding element $R_{w,w}$. The elements of $\bm{R}_k(\psi_0)$ are defined as follows: if the length of the Dubins path to the configuration $\bm{q}_1=[(b - w) l_v, (c-w)l_v, \psi_1]^{\rm T}$, where $b$ and $c$ are the indices for the $x$ and $y$ indices for the reachability matrix, is less than the maximum travel distance $u_{v, {\rm max}}\Delta t$ and greater than the minimum travel distance $u_{v, {\rm min}}\Delta t$, then the corresponding element $({\bm R}_k(\psi_0))_{a,b}=1$ otherwise $({\bm R}_k(\psi_0))_{a,b}=0$. To find the reachability of system \eqref{eq:dynamics} for initial headings $\psi_0 > \pi/2$, the discretized reachability sets in the range $0 \leq \psi_0 < \pi/2$ can be rotated in increments of $\pi / 2$. Since the matrix $\bm{R}_k(\psi_0)$ represents the set of discretized states in the $xy$ plane, the set $\mathcal{R}(\bm{q}_0, k) \subseteq {\rm SE}(2)$ is a set of final configurations, were each element corresponds to one element of $\bm{R}_k(\psi_0)$. The set $\mathcal{R}(\bm{q}_0, k)$ contains all of the possible final positions $q_{1, x}, q_{1, y}$ and final heading angles $\psi_1$ associated with elements of $\bm{R}_k(\psi_0)$.
 
\begin{figure}
    \centering
    \includegraphics{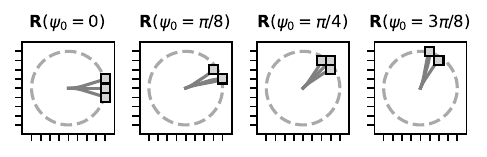}
    \caption{The reachability of the variable-speed Dubins vehicle for different initial headings $\psi_0$.}
    \label{fig:reachability}
\end{figure}

\begin{figure}
    \centering
    \includegraphics{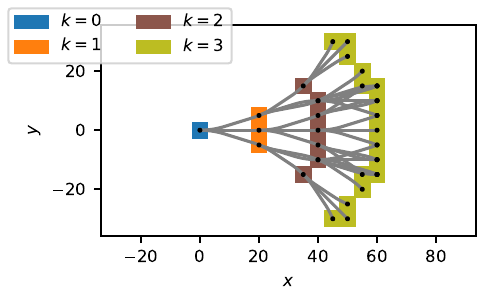}
    \caption{The reachability of the variable speed Dubins vehicle for different timesteps.}
    \label{fig:reachchain}
\end{figure}

An example of the reachability of system \eqref{eq:dynamics} is shown in Fig.~\ref{fig:reachability} with parameters: $u_{\psi, {\rm max}}=\pi / 4$~rad/s, $u_{v, {\rm min}}=18$~m/s, $u_{v, {\rm max}}=22$~m/s, $l_v=5$~m, $\Delta\psi=\pi/8$, and $\Delta t=1$~s. The dark gray lines are the Dubins paths from $\bm{q}_0$ to the voxels with varying heading angle $\psi_0$. The gray dashed line has radius $r=(u_{v, {\rm min}} + u_{v, {\rm max}})\Delta t/2$ for reference.
The Dubins paths from the discretized reachability set for $k=1$ are used as motion primitives and chained together to create the discretized reachability sets for $k>1$. An example of the chained discretized reachability sets is shown in Fig.~\ref{fig:reachchain} with the same set of parameters as Fig.~\ref{fig:reachability}. The initial condition is $\bm{q}=[0, 0, 0]^{\rm T}$. The Dubins paths are shown in gray. The black dots are the $xy$ locations of configurations corresponding to the center of the voxels, shown as multicolored squares.

\subsection{Graph Search}
The $A^\ast$ algorithm with iterative deepening \cite[Ch.~2]{lavalle2006ch2} finds the best path over a time horizon $\tau \in T$, starting at $\bm{q}_k$ and ending at $\bm{q}_{k + \tau}$ for a set of time horizons $T = \{0, \tau_1, \ldots, \tau_{N_h - 1}\}$. Iterative deepening is a horizon planning method for incrementally increasing the time horizon, i.e., $T_1 = \{0, \tau_1\}$, $T_2 =\{0, \tau_1, \tau_2\}$, etc. until $T_{N_h - 1}= \{0, \tau_1, \ldots, \tau_{N_h - 1}\} = T$. In this implementation, the $A^\ast$ algorithm runs for the set of initial time horizons $T_1$ and creates a path $\{\bm{q}_{k}, \bm{q}^1_{k + \tau_1}\}$. Then, the previous $A^\ast$ search is discarded and a new $A^\ast$ search starts with set of horizons $T_2$ finds a new path $\{\bm{q}_{k}, \bm{q}^2_{k + \tau_1}, \bm{q}^2_{k + \tau_2}\}$. Where the superscript represents a unique path for a different time horizon. For problems, such as this one, where there is a large branching factor, the effort of exploring the space with time horizon $T_n$ is negligible compared to the effort of exploring the space with time horizon $T_{n+1}$, hence discarding previously explored paths at earlier time horizons may be justified. Standard iterative deepening proceeds indefinitely, with an increasing time horizon, stopping only when the goal state $\bm{q}_g$ is reached.  In contrast, this work proposes to use a goal node not tied to a state of the system \eqref{eq:dynamics} but rather time. Specifically, at each time horizon, the $A^\ast$ algorithm uses a search graph where all of the nodes at time $\tau_{f}$ are connected to an abstract goal node. The timestep $\tau_f$ starts as $\tau_1$ and increases with each successive run of $A^\ast$ until $\tau_{N_h - 1}$. This modification to the standard $A^\ast$ search graph forces $A^\ast$ to find the best path for the current time horizon, adapting the algorithms for the horizon planning problem \eqref{eq:problem}. Using iterative deepening allows planning to stop execution early, when the entire road network has been searched or after a fixed execution time. When the search problem has been completed, the horizon planner is no longer necessary and terminates early, saving computational resources. The UAV is then free to switch to a different objective, such as maximizing the standoff distance while maintaining visibility \cite{Hague2025}.

\begin{figure*}[t]
    \centering
    \includegraphics{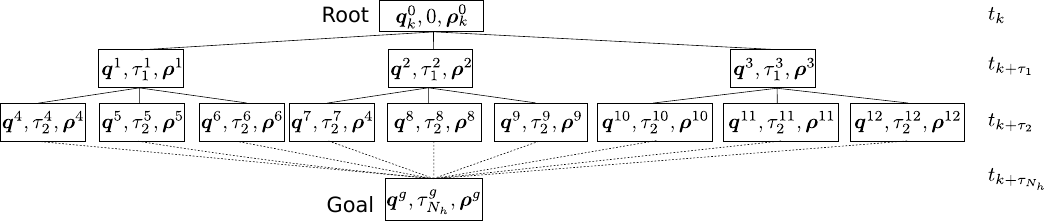}
    \caption{An example search graph that considers two time-steps and a branching factor of $|R_1(\psi_0)|=3$ used to find the path that maximizes the probability of measuring the POI. }
    \label{fig:astar_graph}
\end{figure*}

Let $G_2$ be a hierarchical graph, as shown in Fig.~\ref{fig:astar_graph}, that is defined by a collection of nodes $\mathcal{N}_2$ and edges $\mathcal{E}_2$. The graph $G_2$ is defined for a particular time horizon $T$ within the iterative deepening procedure; the graph $G_2$ is discarded and re-created each time the time horizon deepens and a new A* search is performed. Each node $n_i\in\mathcal{N}_2$ is a tuple consisting of a timestep $\tau^i$, system \eqref{eq:dynamics} state $\bm{q}^i$, and an unobserved probability vector (UPV) $\bm{\rho}^i \in [0, 1]^\delta$. The superscript corresponds to the index of the node $n_i$. Nodes are grouped based on the timestep $\tau^i$, and the groups are arranged in a hierarchy based on the increasing value of $\tau^i$. A UPV is a vector used during path planning whose elements are associated with each node in the target state graph with values representing current (or future) target location probability. However, unlike a typical probability distribution, the UPV sets all values of target probability in nodes that are visible along a UAV's path to zero at the corresponding times. Thus, non-zero values of the UPV indicate regions of the target state graph that are not observed at a given time and the degree to which they \emph{may} contain the POI. The UPV will be determined as the graph $G_2$ is constructed, and evaluating the UPV will leverage \eqref{eq:visibile}, i.e.,  a possible POI state 
$s$ is \emph{unobserved} from state ${\bm q}$ when $f(\bm{q}, s)=0$ and \emph{observed} when $f(\bm{q}, s) = 1$. Given the $a$th element of a UPV vector at time step $k$ (i.e., $\rho_{k, a}$ ),
the UPV vector predicted at $\tau_b$ time steps into the future:
\begin{equation}
    \rho_{k + \tau_b, a} = \rho_{k,a}\prod_{c=0}^{b}\bm{Z}^{\tau_c}(1 - \beta f(\bm{q}_{k + \tau_c}, s_a))\;,
    \label{eq:upv}
\end{equation}
where $0 \leq b \leq N_h -1$ is the index of $\tau_b \in T$, and $\beta \in [0, 1]$ is a constant weighting exploration versus exploitation and can account for the probability of non-detection $1 - p_d$. The sum of the elements of $\bm{\rho}_{k + \tau}$ decreases when the UAV views possible POI states $s \in \mathcal{N}_1$.
When expanding nodes on the graph $G_2$, the probabilities in the UPV corresponding to observed states are modified by multiplying wit the constant value $1 - \beta$ according to \eqref{eq:upv}, while the unobserved states' probabilities in the UPV remain unmodified, multiplied by $1$, but are still propagated forward in time with the motion model. The probability from unobserved states may transition, when propagated forward in time with the POI's motion model $\bm{Z}$, to states that were formally observed but no longer have the UAV in their VVs and are unobserved at the future times considered.  The early stopping condition for the interactive deepening planning is when the UPV is the zero vector indicating that the entire space has been searched.

Now, returning to the construction of $G_2$,  the root node is at the top of the hierarchy and has no parent with the UPV $\bm{\rho}_k^0$ initialized to $p(s|\bm{\xi}_{0:k})$, timestep $t_k$ (the current time when path planning is initiated), and the current state of system \eqref{eq:dynamics} $\bm{q}^0_k$. The subscript on the time horizon $\tau$ denotes the planning timestep, and the node index is denoted with a superscript.  Consider an edge $(n_i, n_j )\in \mathcal{E}_2$ where $n_i \in \mathcal{N}_2$ is a node at timestep $k$ and $n_j \in \mathcal{N}_2$ is a node at timestep $k+\tau$. Edges are directional and only connect nodes in successive timesteps (down the hierarchy) with configurations in the reachability of the current node. In other words, the state ${\bm q}^j \in \mathcal{R}({\bm q}^i, \tau)$ corresponding to node $n_j$ is in the reachability of ${\bm q}^i$ corresponding to node $n_i$ and $t^i + \Delta t (\tau^j - \tau^i) = t^j$.
The relationship between nodes is denoted as parent and child. In the previous $n_i$, $n_j$ example, $n_i$ is the parent node $n_p$ of node $n_j$, while $n_j$ is the child node $n_c$ of node $n_i$. The nodes at the end of an outgoing edge are called children $n_c$, and the node at the start of an outgoing edge is called a parent.
Expanding a parent node involves adding all nodes to reachability as children. The graph $G_2$ has a tree-like structure with a branching factor equal to the cardinality of the discretized reachability for one time step, i.e., $|R_1(\psi_0)|$. However, it differs from a tree since the final node is connected to all nodes at the previous layer and a node can have multiple parents when after expanding parents nodes their children nodes have the same UAV state $\bm{q}$, time $t_{k + \tau}$, and UPV $\bm{\rho}$.

In Fig.~\ref{fig:astar_graph}, each node is represented as a square and connected to other nodes with directional edges, solid lines. The transition cost associated with the edge $(n_i, n_j)$  will be described shortly (in \eqref{eq:transition}) and is based on the states visible from the UAV, i.e., those states $s$ that satisfy $f(\bm{q}_{k + \tau}, s) = 1$, and the value of the UPV vector ${\bm \rho}_{k + \tau}$ with elements corresponding to the states $s$.
The dashed lines represent zero-cost edges connecting nodes at the final timestep to the goal node, which terminates the $A^\ast$ search.
The graph $G_2$ differs from prior work \cite[Ch.~2]{lavalle2006ch2} because the nodes in $G_2$ have a temporal structure and encode for the UPV in addition to encoding the state of the system \eqref{eq:dynamics}. 

The maximization \eqref{eq:problem} can be broken into stages that represent the weights of edges shown as solid black lines on the graph in Fig.~\ref{fig:astar_graph}. The $A^{\ast}$ algorithm minimizes the total cost of traveling to different nodes on a weighted graph. First, it is necessary to convert the maximization \eqref{eq:problem} into a minimization by defining the cost to transition from a node with configuration $\bm{q}^p \in {\rm SE}(2)$, timestep $\tau^p$, and UPV $\bm{\rho}^p$ to a node with configuration $\bm{q}^c\in {\rm SE}(2)$, timestep $\tau^c$, and UPV $\bm{\rho}^c$. The cost of transitioning from one node to another is approximately one element of the summation in \eqref{eq:problem}. That is, the transition cost is the discounted probability of not viewing the POI,
\begin{equation}
    \label{eq:transition}
    c(\bm{q}^{c}, \bm{\rho}^{c})  = 1 - \gamma^{\tau^{c}}\sum_{j=0}^{\delta - 1} f(\bm{q}^c, \bm{g}_{s,j})\rho^c_j
\end{equation}
where $\bm{g}_{s, j}$ is the planar position of state $s_j$. Equation \eqref{eq:transition} is obtained by taking the negative of \eqref{eq:prob_detection} in \eqref{eq:problem} and assuming $\mu = 0$ and $p_d = 1$ (i.e., a perfect sensor). The assumption of a perfect sensor for \eqref{eq:transition} is justified because it would uniformly increase the cost for all nodes if the actual constants for $\mu$ and $p_d$ are used. Adding one ensures the costs are positive numbers and avoids negative edge weights. Negative edge weights violate the assumptions of the $A^*$ algorithm. Moreover, where the actual cost \eqref{eq:problem} is based on the actual future distribution $p(s_i| \bm{\xi}_{0:k-1})$, the cost \eqref{eq:transition} for path planning uses the UPV $\rho^c_j$, which is an approximation of this distribution.

When creating a child node on $G_2$, its UPV is updated with the parent's UPV $\bm{\rho}^p$ according to two equations, broken out from \eqref{eq:upv}. 
The first is \emph{observing} possible POI locations $s_i$ from the UAV's location $\bm{q}^p$ and updating the associated UPV elements $(\rho^p_i)'$,
\begin{equation}
    (\rho^p_i)' = \left[1 - \beta f(\bm{q}^p, \bm{g}_{s,i})\right]\rho^p_i\;,
    \label{eq:collect}
\end{equation}
where $f(\bm{q}^p, \bm{g}_{s,i})$ is \eqref{eq:visibile} and describes the visibility of the $i$th POI state $s_i$ at planar position $\bm{g}_{s, i}$ from the UAV's state $\bm{q}^p$. Equation \eqref{eq:collect} determines what visibility volumes that the UAV is inside at $\bm{q}^p$. The unobserved probabilities, corresponding to the visibility volumes the UAV is inside, are removed from the updated UPV $\bm{\rho}_p$ by assuming the UAV can measure the new states without non-detections and false alarms. If the position of the child node $\bm{q}^c$ was considered and $\beta=1$, then the cost of transition to the child node using \eqref{eq:transition} would always be zero because all of the states visible from $\bm{q}^c$ would be observed.
The second equation \eqref{eq:reward_prop} propagates the unobserved probability forward in time from the parent's planning timestep $\tau^p$ to the child's planning timestep $\tau^c$ with the matrix representation $\bm{Z}$ of the POI's motion model \eqref{eq:markov}:
\begin{equation}
    \bm{\rho}^c =\bm{Z}^{\tau^c - \tau^p}(\bm{\rho}^p)' \label{eq:reward_prop}\;.
\end{equation}
When expanding new nodes on the graph using the discretized reachability, a new node may duplicate an existing one. Nodes are considered the same if the UAV's state, timestep, and UPV are the same, and in such cases, $A^\ast$ may update the cost to travel to the existing node if the new cost to travel to the node is less than the previous cost to travel to.

\subsection{Heuristic Cost to Goal}

\begin{figure}
    \centering
    \includegraphics[width=3.25in]{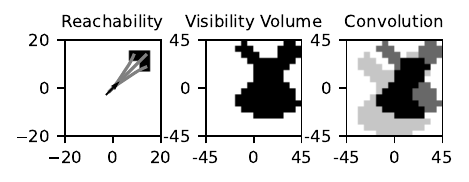}
    \caption{The reachability of a variable-speed Dubins vehicle convolved with a VV and saturated between 0 and 1.}
    \label{fig:reach_convolution}
\end{figure}

The $A^{\ast}$ algorithm uses a heuristic lower bound on the cost of traveling from the current state to the goal to guide the search. This work introduces a heuristic \eqref{eq:heuristic_matrix} that estimates the amount of unobserved probability viewable by system \eqref{eq:dynamics},
if system \eqref{eq:dynamics} could visit all of the voxels in its reachability at the same time. This promotes visiting a promising state during the search that has an increased likelihood of viewing the POI in the future. The cost of traveling to the goal node is approximated by a matrix of elements that accounts for the cost function \eqref{eq:transition}, the discretized reachability $\bm{R}$, and the VVs $\bm{\mathfrak{a}}$:
\begin{multline}
    \bm{\mathfrak{R}}(\bm{\rho}_{k + \tau_a}) = \sum_{i=a + 1}^{N_h}\Bigg\{1 -\gamma^{\tau_i}\sum_{j=0}^{\delta -1}\bigg[\left(\bm{Z}^{\tau_i - \tau_a}\bm{\rho}_{k + \tau_a}\right)_j\\{\texttt{sat}}\left(\bm{R}_{\tau_i - \tau_a}\ast \bm{\mathfrak{a}}(\bm{g}_{s_j}), 0, 1\right)\bigg]\Bigg\}\;,
    \label{eq:heuristic_matrix}
\end{multline}
where  $\left(\bm{Z}^{\tau_i - \tau_a}\bm{\rho}_{k}\right)_j$ is the $j$th element of the UPV propagated forward $\tau_i - \tau_a$ timesteps with the POI's motion model, $\texttt{sat}(\bm{A}, a, b)$ is the saturation of the elements of the matrix $\bm{A}$ to between $a$ and $b$, and $(\cdot) \ast (\cdot)$ is the matrix convolution operation. The heuristic is obtained by considering the sum of probabilities associated with the reachable VVs. The summation in \eqref{eq:transition} is mirrored in \eqref{eq:heuristic_matrix} with the inner summation. The visible probability term $f(\bm{q}_c, \bm{g}_{s, j})\rho_{c,j}$ is approximated for all possible $\bm{q}$ by summing the VV matrix elements multiplied by the UPV associated with the VVs for a single instance in time. The outer summation considers the other timesteps in the time horizon. In the inner summation, the UPV is propagated forward in time $\tau_i - \tau_a$ timesteps, and each element of the resulting vector is multiplied by the convolved and saturated VV. Each element of the resulting vector is the unobserved probability that the UAV could view the node in $\tau_i - \tau_a$ timesteps. The convolution is saturated between 0 and 1 because a VV is a binary scalar field accounting for set membership. If the inertial component of a UAV configuration $[q_x, q_y, h_{\rm UAV}]^{\rm T} \in \alpha(\bm{g})$ then the scalar field is one, otherwise zero. The convolved VV is also a binary scalar field with the condition ``From what initial conditions can the UAV reach the VV in $\tau_i - \tau_a$ steps while accounting for the motion of the POI ?''. The matrix convolution is a map $\ast : \mathbb{R}^{a_0\times a_1}\times \mathbb{R}^{b_0\times b_1}\rightarrow\mathbb{R}^{b_0\times b_1}$, where $a_0 \leq b_0$, $a_1 \leq b_1$ and $a_0, a_1$ are odd, and can be written as
\begin{equation}
    (\bm{A} \ast \bm{B})_{x,y} = \sum_{i=-\lfloor a_0 / 2\rfloor}^{-\lfloor a_0 / 2\rfloor}\sum_{j=-\lfloor a_1 / 2\rfloor}^{-\lfloor a_1 / 2\rfloor} A_{i,j}B_{x+i, y+j}\;.
\end{equation}
If indices to the matrix $B$ are invalid $x+i < 0$, $x + i \geq b_0$, $y+j < 0$, or $y+j \geq b_1$ then a value of $0$ is used for $B_{x + i, y + j}$. 
A convolved and saturated VV, henceforth called a future reachable VV state set (FRVVSS), is the matrix ${\texttt{sat}}\left(\bm{R}_{\tau_i - \tau_a}\ast \bm{\mathfrak{a}}(\bm{g}_{s_j}), 0, 1\right)$ from \eqref{eq:heuristic_matrix}, where if system \eqref{eq:dynamics} starts at a particular voxel, given an initial heading, the associated element of the matrix is one when at least one voxel of the VV is in the reachability and zero when no voxels of the VV are in the reachability. 
During execution of the $A^\ast$ algorithm, the heuristic cost to goal from a particular configuration $\bm{q}_{k + \tau}$ is an element of \eqref{eq:heuristic_matrix},
\begin{equation}
    r(\bm{q}_{k + \tau}, \bm{\rho}_{k + \tau})=\bm{\mathfrak{R}}(\bm{\rho}_{k + \tau})_{ij}\;,
    \label{eq:heuristic}
\end{equation}
where $i$ and $j$ are the indices of a voxel where the center of the voxel is $[q_{k + \tau, x}, q_{k + \tau, y}, h_{\rm UAV}]^{\rm T}$.
An example of \eqref{eq:heuristic_matrix} is shown in Fig.~\ref{fig:reach_convolution}. 
The heuristic matrix \eqref{eq:heuristic_matrix} is the sum at each timestep of one minus the discounted sum of the probability associated with the FRVVSS. The heuristic is an underestimate for the cost of traveling to a particular voxel because, at best, system \eqref{eq:dynamics} can travel to all of the VVs (each VV is associated with a state $s\in N_1$). At worst, system \eqref{eq:dynamics} travels to none of the VVs and has a cost higher than the heuristic. 
In the left panel of Fig.~\ref{fig:reach_convolution}, the reachability of a variable-speed Dubins vehicle with initial conditions $\bm{q}_0 = [0, 0, \pi/4]^{\rm T}$, altitude $h_{\rm UAV}=75$~m, and parameters from Fig.~\ref{fig:reachability}. The middle panel represents a single VV as a matrix $\bm{\alpha}(\bm{g})$ at altitude $h_{\rm UAV} = 75$~m and sensing distance $l_{\rm max}=300$~m, where the black pixels are ones and the white pixels are zeros. The right panel shows the FRVVSS and the VV. The light gray is the FRVVSS, the dark gray is the VV, and the black is the overlap of the two sets.

\subsection{Variable-Timestep Planning}
\begin{figure}[t]
    \centering
    \includegraphics[width=3.25in]{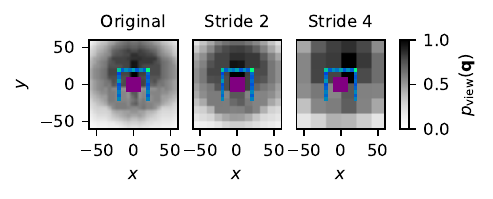}
    \caption{The probabilistic visibility volume for different max-pooling stride lengths.}
    \label{fig:maxpool}
\end{figure}

To plan further into the future, the UAV's configuration space can be reduced by max-pooling the probabilistic VVs. Max-pooling is the operation that reduces the size matrix by breaking the matrix into subsections and reducing the subsection to its maximum  value \cite{scherer2010}. The subsection of the matrix is a square matrix where the stride length is the number of rows and columns in the subsection. The equation for max-pooling a matrix $\bm{A} \in \mathbb{R}^{a \times a}$ with stride length $b$ is
\begin{equation}
    B_{x,y} = \max\left(\begin{bmatrix}
        A_{x,y} &\cdots&A_{x+b-1,y}\\
        \vdots&\ddots&\vdots\\
        A_{x, y+b-1} & \cdots & A_{x+b-1, y+b-1}
    \end{bmatrix}\right)\;,
\end{equation}
where $B_{x,y}$ is the element of the resulting matrix $\bm{B}$ which has size $\lceil\frac{a}{b}\rceil\times\lceil\frac{a}{b}\rceil$.
By using a stride length of two, the size of a voxel doubles and decreases the probabilistic VVs' resolution. System \eqref{eq:dynamics} can efficiently plan over a coarse FRVVSS further into the future for the same computational cost. Increasing the size of the voxels reduces the number of outgoing edges on each node and decreases the depth of the graph $G_2$. For graphs with the same time horizon, increasing the size of the voxels decreases the depth of the graph and reduces the time needed to search the graph for an optimal path. Figure~\ref{fig:maxpool} shows the probabilistic VV at an altitude $h_{\rm UAV}$ for a simple ``U" shaped road network. The probability of viewing the POI is shown as the white to black color scale. The probability distribution of the POI on the road network is shown with the blue to green color scale. The purple square in the middle is the occluding obstacle.
When the path planner expands a node with reduced resolution, \eqref{eq:collect} is used on all of the possible configurations contained in the reduced resolution voxel.

\subsection{Planning Algorithm}
The path planning algorithm for searching for a moving POI is broken into three parts: iterative deepening (Algorithm~\ref{alg:iterative}), $A^{\ast}$ (Algorithm~\ref{alg:astar}), and the tree node expansion procedure (Algorithm~\ref{alg:neighbors}). Algorithm~\ref{alg:iterative} is a standard implementation with additional stopping conditions, and Algorithm~\ref{alg:astar} is a standard implementation that uses the heuristic \eqref{eq:heuristic} and Algorithm~\ref{alg:neighbors} to expand new nodes. Both Algorithms~\ref{alg:iterative}--\ref{alg:astar} are described in the Appendix, Sec.~\ref{sec:appendix}. Here we describe Algorithm~\ref{alg:neighbors}, which is a novel way of expanding nodes to create the graph structure $G_2$ in Fig.~\ref{fig:astar_graph}.

Algorithm~\ref{alg:neighbors} takes a parent configuration $\bm{q}_{k + \tau_p}$, a UPV $\bm{\rho}_{k + \tau_p}$, a timestep $t_{k + \tau_p}$, and a set of timesteps $T$ and returns a set of neighboring configurations, UPVs, and the next timestep that are added to the graph. The algorithm starts by setting the set of neighbors $\omega$ to an empty set (line~\ref{alg:neighbors:init}). Then, the algorithm checks to see if the goal node---a node with no state connected to all of the nodes at the final timestep---is next (line~\ref{alg:neighbors:isGoal0}). If the goal node is next, the set of neighbors is returned as only the goal node, where $-1$ is not an element of the set $T$ (line~\ref{alg:neighbors:retgoal}). Next, the parents's intermediate UPV is calculated using \eqref{eq:collect} (lines~\ref{alg:neighbors:collect0}--\ref{alg:neighbors:collect1}), and the parent's intermediate UPV is propagated to the childrens' timestep \eqref{eq:reward_prop} (line~\ref{alg:neighbors:propigate}). Finally, all of the states reachable from the parent node $\bm{q}_{p}$ in $\tau_{p + 1} - \tau_p$ timesteps are added to the set of neighbors $\omega$, with UPV $\bm{\rho}_c$ and timestep $\tau_{p + 1}$, and returned (lines~\ref{alg:neighbors:reach0}--\ref{alg:neighbors:return}). If child node duplicates an already existing node, then the $A^\ast$ algorithm handles the duplicate.
\begin{algorithm}[h!]
  \caption{Neighbors}
  \begin{algorithmic}[1]
    \STATEx \hspace{-2.1em}  {\bf function:} $\texttt{Neighbors}(\bm{q}_{k + \tau_p},\;\bm{\rho}_{k + \tau_p},\;\tau_p,\;\tau_{p + 1},\;T, \bm{Z}, \beta)$
    \STATEx \hspace{-2.1em}   {\bf input:} $\bm{q}_{k + \tau_p}$ the parent's position, $\bm{\rho}_{k + \tau_p}$ the parent's reward vector, $\tau_{p}$, the parent's planning timestep, $\tau_{p + 1}$ the children node's timestep, $T$ the set of time horizons for the horizon planner. $\bm{Z}$ the POI's motion model, $\beta$ the constant weighting exploration versus exploitation.
    \STATEx \hspace{-2.1em}  {\bf output:} $\omega = \{(\bm{q},\tau,\bm{\rho})\}$ a set of neighbors
    \STATE $\omega \gets \emptyset$ // Initialize set of neighbors \label{alg:neighbors:init}
    \IF{$\tau_{p + 1}\notin T$} \label{alg:neighbors:isGoal0}
        \STATE $\RETURN \{(\emptyset, -1, \bm{\rho}_{k + \tau_p})\}$ \label{alg:neighbors:retgoal}
    \ENDIF \label{alg:neighbors:isGoal1}
    \FOR{$s_i \in \mathcal{N}_1$} \label{alg:neighbors:collect0}
        \STATE $\rho'_{k + \tau_p,i} \gets[1- \beta f(\bm{q}_{k + \tau_p}, \bm{g}_{s,i})]\rho_{k + \tau_p,i}$ // Visibility update of UPV
    \ENDFOR \label{alg:neighbors:collect1}
    \STATE $\bm{\rho}_c \gets \bm{Z}^{\tau_{p + 1} - \tau_p}\bm{\rho}'_{k + \tau_p}$ // Motion update of UPV \label{alg:neighbors:propigate}
    \FOR{$\bm{q}_c \in \mathcal{R}(\bm{q}_{k + \tau_p}, \tau_{p + 1} - \tau_p)$} \label{alg:neighbors:reach0}
        \STATE $\omega \gets \omega \cup \{(\bm{q}_c, \tau_{p + 1}, \bm{\rho}_c)\}$ // Add states in reachibility
    \ENDFOR\label{alg:neighbors:reach1}
    \STATE $\RETURN \omega$ \label{alg:neighbors:return}
  \end{algorithmic}
  \label{alg:neighbors}
\end{algorithm}

The overall planning approach starts by using a map of the environment and road network. All visibility volumes and the UAV's reachability are pre-calculated for the planning algorithm to increase the planner's speed. The UAV assumes an initial distribution of possible POI states. First, the UAV plans its path with the POI probability distribution using ID $A^\ast$. Second, the UAV selects the path with the longest time horizon and moves forward. Third, the UAV collects a measurement and uses the RBE to estimate the state of the POI. The UAV loops the three steps until the POI is found, completing the search.

%% file: sections/05_Simulation.tex
\begin{figure}
    \centering
    \includegraphics{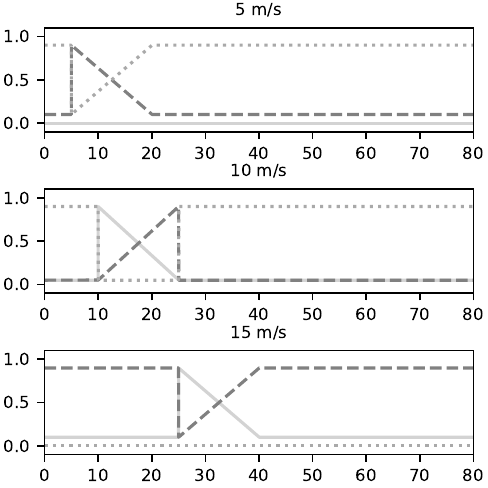}
    \caption{The POI's probability of changing speed based on the distance to an intersection and the current speed.}
    \label{fig:speed}
\end{figure}

This section presents two examples that help illustrate the approach and a Monte-Carlo study that evaluates the proposed approach compared to a baseline static VV-based coverage planner \cite{Hague2023} and a lawnmower coverage planner.\footnote{The implementation of this study can be found at \href{https://github.com/robotics-uncc/DubinsVisibilitySearch}{https://github.com/robotics-uncc/DubinsVisibilitySearch}} 
To evaluate the estimation of the POI, this work uses the variance of a graph from \cite{Coscia2021, Devriendt2022},
\begin{equation}
    \sigma_n^2=\sum_{n_i, n_j\in \mathcal{N}_1}p(s_i, \xi_{0:k})p(s_j, \xi_{0:k})d_n(\bm{g}_{s_i}, \bm{g}_{s_j})^2\;,
\end{equation}
where $\sigma_n^2$ refers to the position variance, $p(s_i, \xi_{0:k})$ is the probability that the POI is at the $i$th state, and $d_n(\bm{g}_{s,j}, \bm{g}_{s,j})$ is the distance in the $xy$ plane along the graph between $s_i$ and $s_j$ calculated with the Floyd-Warshall algorithm \cite[Ch.~25]{Cormen2009}, where the edge weights are the Euclidean distance between the nodes. The POI's covariance on the graph $G_1$ is
\begin{equation}
    \bm{P}=\begin{bmatrix}
        \sigma_n^2 & 0 \\
        0 & \sigma_v^2
    \end{bmatrix}\;,
    \label{eq:poicovar}
\end{equation}
where
\begin{equation}
    \sigma_v^2=\sum_{n_i, n_j\in \mathcal{N}_1}p(s_i, \bm{\xi}_{0:k})p(s_j, \bm{\xi}_{0:k})(v_{s_j} - v_{s,i})^2\;,
\end{equation}
is the speed variance.
In the numerical study, the POI's discrete Markov chain, with a maximum speed of $15$~m/s, has the turn transition probabilities at intersections from Table~\ref{tbl:intersection} (the intersection labels are shown in Fig.~\ref{fig:cells}), and the speed transition probabilities from
\begin{table}[t]
    \centering
    \small
    \caption{The POI's probability of going in any direction at an intersection based on the current speed.}
    \begin{tabular}{cccc}
        \hline\hline
        & 5 (m/s) & 10 (m/s) & 15 (m/s) \\
        \hline
        $\epsilon_0$ & 0.05 & 0.0 & 0.0\\
        $\epsilon_1$ & 0.225 & 0.15 & 0.075\\
        $\epsilon_2$ & 0.5 & 0.7 & 0.85 \\
        $\epsilon_3$ & 0.225 & 0.15 & 0.075 \\
        $\epsilon_4$ & 0.05 & 0.0 & 0.0 \\
        $\epsilon_5$ & 0.475 & 0.5 & 0.5 \\
        $\epsilon_6$ & 0.475 & 0.5 & 0.5 \\
        $\epsilon_7$ & 0.05 & 0.0 & 0.0 \\
        $\epsilon_8$ & 0.575 & 0.75 & 0.875 \\
        $\epsilon_9$ & 0.375 & 0.25 & 0.125 \\
        
        \hline
    \end{tabular}
    \label{tbl:intersection}
\end{table}
Fig.~\ref{fig:speed}. The motion model approximates a car slowing down to make turns at intersections. 
In this study, road network node spacing is $l_c=5$~m, and the timestep is $\Delta t=1$~s. At each timestep, the UAV uses iterative deepening to plan multiple paths and uses the first control from the longest path plan each timestep.

\subsection{Illustrative Examples}
\begin{figure}
    \centering
    \includegraphics{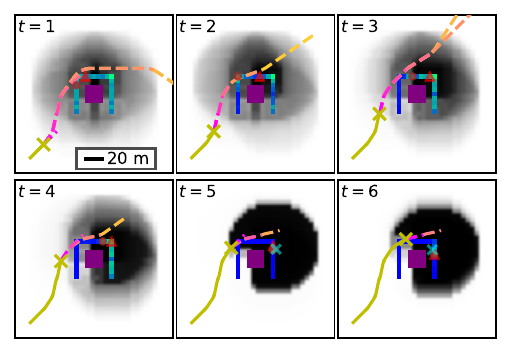}\\
    \includegraphics{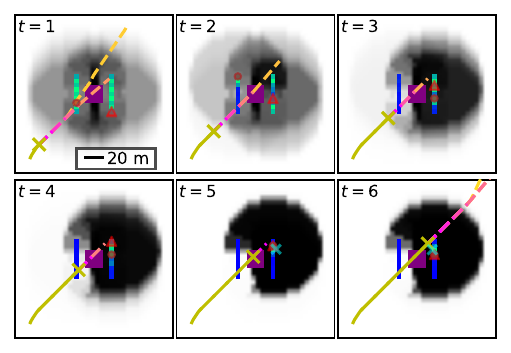}
    \caption{The UAV searching for a target on a ``U'' shaped network (top) and a parallel road network (bottom) with an occlusion in the middle.}
    \label{fig:uticker}
\end{figure}

Two examples of the search method are shown with (i) an environment with a ``U'' shaped road with an obstacle in the middle, and (ii) an environment with two parallel roads with an obstacle in the middle. The initial probability distribution is a uniform weighting over the road network. The iterative deepening planner uses the time horizon $T=\{1, 2, 3, 5, 7, 9, 13\}$ and timestep $\Delta t=1$~s. An example of the proposed planner is shown in Fig.~\ref{fig:uticker}. The simulations use initial condition $\bm{q}_0 = [-75,  -75, \pi/2]^{\rm T}$, maximum speed $22$~m/s, minimum speed $18$~m/s, maximum turn-rate control $u_{\psi, {\rm max}}=\pi/4$, probability of detection of $p_d=1$, error probability $\mu=0$, measurement covariance $$
\bm{R} = \begin{bmatrix}
    20 & 0\\
    0 & 20
\end{bmatrix}\;,
$$ maximum POI speed $15$~m/s, $l_v=5$~m, environment bounds $x_{\rm min}=-100,~x_{\rm max}=100,~y_{\rm min}=-100,~y_{\rm max}=100$, exploration factor $\beta = 1$, and discount factor $\gamma=0.1$. The choice of $p_d = 1$, $\mu = 0$, and $\beta = 1$ are to simplify the example for illustrative purposes. The probability distribution is shown as the blue to green color map, normalized so green is the maximum probability at each timestep. The white to black color map represents the increasing probability that the UAV can view the POI. The solid yellow line is the UAV's past path, while the dashed lines are the paths found by the iterative deepening planner. The color map, pink to yellow, shows an increasing planning horizon. The yellow ``x'' is the UAV's position, the cyan ``x''s are the UAV's measurements of the POI, the red triangle is the POI's position, and the brown circle is the mode of the UAV's estimate. The timestep is shown on each subplot.

The top example of Fig.~\ref{fig:uticker} shows the UAV flying towards the region where the UAV is most likely to measure the POI for $t=1$~s to $t=4$~s. When UAV measures the POI for the first time at $t=5$~s, reducing the covariance, the iterative deepening method reaches the early stopping condition after planning three timesteps into the future. The search stopped because the UAV observed all of the probabilities in the UPV after three timesteps. This results in the UPV being the zero vector, a condition that signifies the end of the search. Once the UAV has an accurate estimate of the POI's state, the UAV can switch to a tracking method (e.g. \cite{Hague2025}) to maintain visibility of the POI.

In the bottom example of Fig.~\ref{fig:uticker}, the UAV flies towards the location with the highest probability of viewing the POI, the middle of the environment above the purple obstacle. Once the UAV has measured the POI in $t=3$~s and $t=4$~s, the UAV has searched the entire left road segment, and estimated that the POI is on the right road segment. Then, at $t=5$~s the UAV measures the POI for the first time, completing the search, and can switch to a tracking method to maintain visibility.

\subsection{Monte Carlo Simulation}
\begin{figure}
    \centering
    \includegraphics{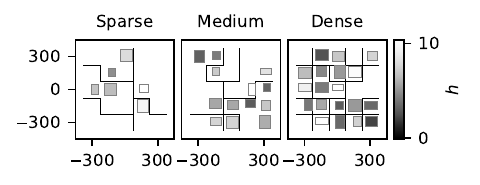}
    \caption{Sparse, medium, and dense environments to evaluate the proposed algorithm.}
    \label{fig:monte_envs}
\end{figure}

A Monte Carlo simulation evaluates the planner on three different environments: sparse, medium, and dense, shown in Fig.~\ref{fig:monte_envs}. 
The environments were created using the wave function collapse (WFC) algorithm \cite{WFC2016} with five road network tiles with side length $150$~m: straight, turn, fork, four-way, and empty. The weights of the different tile types are shown in Table~\ref{tbl:weights}. The WFC generator is run until the road network is fully connected. The buildings, extruded rectangles, are randomly placed in grid cells between the roads with the probabilities given in the right column of Table~\ref{tbl:weights}. Each building is at least $15$~m from the edge of the grid cell or the center of a possible road. The building's side lengths are randomly generated between $60$~m and $120$~m, and heights are randomly generated between $10$~m and $50$~m. The bounds on the environment are $x_{\rm min}=-450,~x_{\rm max}=450,~y_{\rm min}=-450,~y_{\rm max}=450$.

\begin{table}[]
    \centering
    \caption{Wave Function Collapse Weighting for generating the Monte Carlo environments.}
    \footnotesize
    \begin{tabular}{ccccccc}
         \hline\hline
         & Straight & Turn & Fork & Four-way & Empty & Building \\
         \hline
         Sparse & 0.17 & 0.17 & 0.087 & 0.043 & 0.52 & 0.3\\
         Medium & 0.21 & 0.21 & 0.21 & 0.16 & 0.21 & 0.5\\
         Dense & 0.13 & 0.10 & 0.21 & 0.52 & 0.042 & 0.7\\
         \hline
    \end{tabular}
    \caption*{}
    \label{tbl:weights}
\end{table}
The simulations run with fixed parameters: maximum turn-rate $u_{\psi, {\rm max}}=\pi / 4$~rad/s, probability of detection $p_d=0.8$, exploration factor $\beta = 1$ to weight for exploration for the search problem, reward discount factor $\gamma=0.1$, initial heading $q_{\psi, 0} = \pi / 4$, max sensing distance $l_{\rm max}=300$~m, min speed $u_{v, {\rm min}}=36$~m/s, max speed $u_{v, {\rm max}}=44$~m/s, voxel length $l_v=10$~m, and starts in the voxel containing the $xy$ pair $[-350, -350]^{\rm T}$.
The proposed method runs every timestep until the iterative deepening algorithm plans paths for all time horizons $T=\{1, 2, 3, 5, 7, 9, 13\}$, or the planner executes for $\lambda = 5$~s. The FRVVSS are max-pooled with stride length 2 between timesteps $\tau_3=3$ and $\tau_4=5$, $\tau_4=5$ and $\tau_5=7$, and $\tau_5 = 7$ and $\tau_6 = 9$. The FRVVSS are max-pooled by 4 between timesteps $\tau_6=9$ and $\tau_7=13$.
A baseline method from \cite{Hague2023} is used for comparison. The approach in \cite{Hague2023} is adapted to create an efficient search path that considers occlusions by solving the Dubins traveling salesperson problem with neighborhoods (DTSPN) using the intersecting 3D edge sampling method described in \cite{Hague2023} with $6$ samples per visibility volume and $3$ headings per sample. Each visibility volume is placed in a uniform grid spaced $150$~m apart if a corresponding road network node exists. The result of this is a tour that efficiently covers all the road network considering the visibility of multiple road network segments from a single vantage point and the presence of occlusions.

Another baseline method for comparison is a lawnmower coverage path that flies directly over the road network in the vertical direction. The spacing for the lawnmower sweep is $150$~m. A larger sweep width may be more efficient for non-urban environments, but a flight path directly over the roads is guaranteed to view all of the search space. The UAV starts at the same initial condition as the proposed and DTSPN methods, travels to the start of the lawnmower path, and completes the lawnmower path.
In the baseline methods, the UAV files at nominal speed  $v = \frac{1}{2}(u_{v, {\rm max}} + u_{v, {\rm min}})$ and turns with radius $r_{\rm min} = v/u_{\psi, {\rm max}}$.
The proposed and baseline methods are evaluated over combinations of different false alarm probabilities ($\mu=0.164$, $\mu=0.268$, and $\mu=0.329$), and environments, shown in Fig.~\ref{fig:monte_envs}. Each set of parameters is evaluated with 50 different random POI initial conditions. The proposed planner and lawnmower paths run for 120~s. The DTSPN runs for the time it takes to fly to the tour and complete two tour orbits, which is less than 120~s for all environments. The simulations are terminated early, when the POI is considered localized (or found) when the trace of the covariance \eqref{eq:poicovar} is ${\rm Tr}(\bm{P}) \leq 5$.

\begin{figure}
    \centering
    \begin{subfigure}[b]{3.0in}
        \centering
        \includegraphics[width=\textwidth]{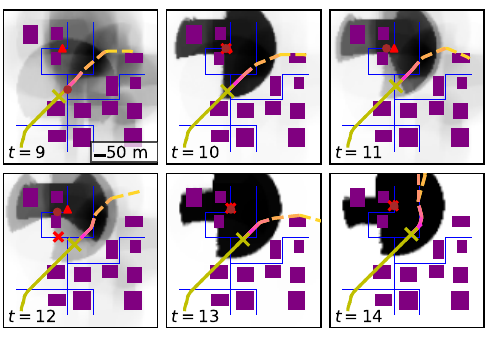}
        \caption{ID $A^\ast$ (Proposed approach)}
    \end{subfigure}\\
    \begin{subfigure}[b]{3.0in}
        \centering
    \includegraphics[width=\textwidth]{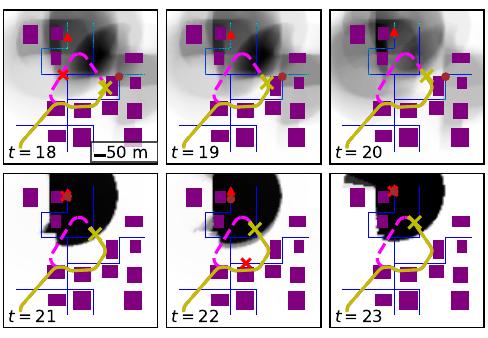}
        \caption{DTSPN (Comparison approach)}
    \end{subfigure}\\
    \begin{subfigure}[b]{3.0in}
        \centering
    \includegraphics[width=\textwidth]{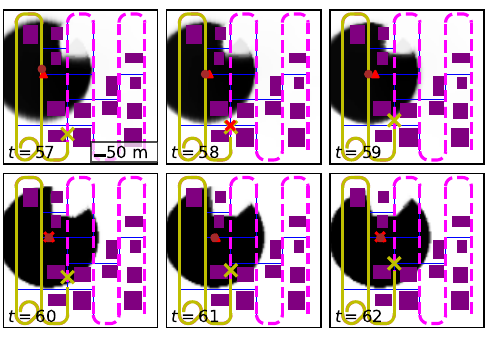}
        \caption{Lawnmower (Comparison approach)}
    \end{subfigure}
    \caption{The proposed method searching for the target (ID $A^*$), the Dubins traveling salesperson problem with neighborhoods (DTSPN) method, and the lawnmower method.}
    \label{fig:examples}
\end{figure}

Three examples of Monte-Carlo runs are shown in Fig.~\ref{fig:examples}, where the error probability is $\mu=0.164$. The top run shows the proposed method, where the yellow line and ``x'' show the UAV's path and current position. The white to back color shows the increasing probability of viewing the POI, and the blue to green color along the roads shows the POI's estimated position. The red triangle is the true POI position, and the brown circle shows the POI's estimated position. The red ``x''s are measurements. The purple rectangles are buildings that can obscure the UAV's view of the POI. The dashed lines, pink to yellow, show the paths with different time horizons resulting from iterative deepening. The middle plot shows the baseline VV-based static point coverage planner (the DTSPN) in magenta. The bottom plot shows the baseline lawnmower coverage path plan in magenta. In the lawnmower example, the POI was measured on the second vertical path, but the trace of the POI's state covariance is not below the cutoff at $t=57$~s. On the third vertical path, the POI is measured again and the covariance is less than the cutoff. In these examples, the proposed method finds the POI in $11$~s while the baseline VV-method finds the POI in $24$~s and the baseline lawnmower method finds the POI in $62$~s.

\begin{figure}[h]
    \centering
    \includegraphics[width=3.25in]{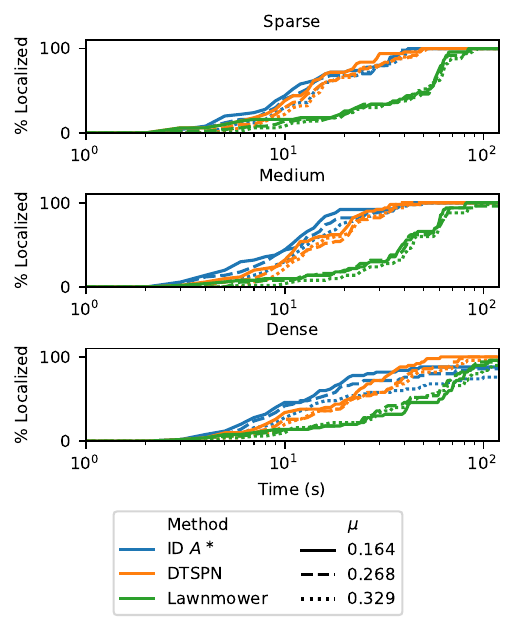}
    \caption{A plot of the percentage of localized POIs for different error probabilities and environments.}
    \label{fig:numLocal}
\end{figure}

In Figure~\ref{fig:numLocal}, the percentage of POIs localized over the logarithm of time is plotted for the different methods and error probabilities. The proposed method, iterative deepening $A^\ast$ (ID $A^\ast$), is shown in blue, the baseline Dubins traveling salesman problem with neighborhoods (DTSPN) is shown in orange, and the lawnmower method is shown in green. The different error probabilities are shown as different line styles: $\mu=0.164$ is solid, $\mu=0.268$ is dashed, and $\mu=0.329$ is dotted. From the graphs, the lawnmower is slower in localizing the POIs for all error probabilities. The ID $A^\ast$ and DTSPN methods perform similarly in the sparse environment, regardless of error probability. The increasing probability of false alarms increases the time it takes to localize the POIs. The styled lines are below the solid ones. The lawnmower and proposed methods fail to localize all of the POIs in the dense environment. The lawnmower method was observed to fail because it moves inefficiently and does not double back to revisit already observed areas. The ID $A^\ast$ method failed to localize all of the POIs because it greedily moves through the middle of the environment, maximizing the chance of viewing the POI, then fails to create a path as efficient as the DTSPN solution at covering the rest of the space. The proposed method may create paths as efficient as the DTSPN when given larger time horizons and more computational resources. The UAV in the DTSPN method flies to the tour from the initial position and flies around the tour twice. Table~\ref{tbl:one_tour} shows the percentage of POI localized after one tour of the environment.
\begin{table}[h]
    \centering
    \small
    \caption{Percentage of POIs localized}
    \label{tbl:one_tour}
    \begin{tabular}{cccc}
        \hline\hline
        Environment & $\mu =0.164$  & $\mu=0.268$ & $\mu=0.329$ \\
        \hline
        Sparse & 94\% & 88\% & 90\% \\
        Medium & 94\% & 88\% & 88\% \\
        Dense & 92\% & 84\% & 80\% \\
        \hline
    \end{tabular}
    \caption*{The percentage of POIs localized after one DTSPN tour.}
\end{table}
The table suggests that a single coverage tour is insufficient for a search problem with false alarms because the POI can move from the unsearched region to the searched region while the POI is outside of the UAV's sensing region.

\begin{figure}[h]
    \centering
    \includegraphics[width=3.25in]{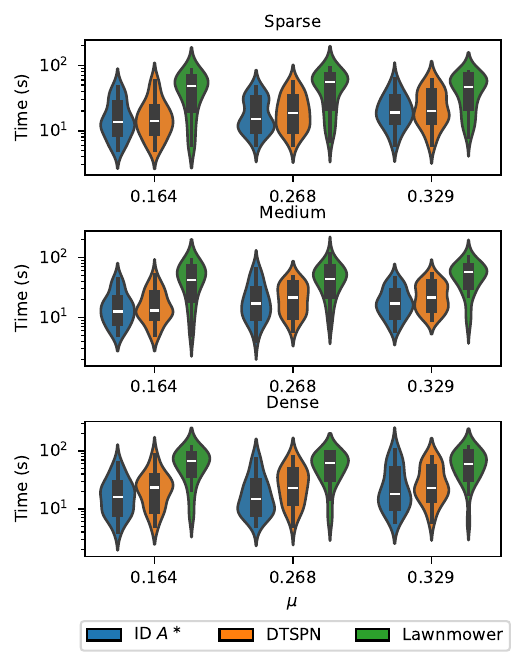}
    \caption{A plot of the time it took each planning method to localize the POI for different error probabilities and environments.}
    \label{fig:findTimes}
\end{figure}

Figure~\ref{fig:findTimes} shows violin and box plots of the time it took to localize the POIs for different environments and error probabilities. The violin plot shows the approximate density of the data points with the varying colored regions. The box plot in black shows the range of the data with the solid black line, the middle 50\% quartile with the black box, and the median of the data with the white line. Data points are included in a set of plots when all methods localize the POI. For example, the lawnmower method does not localize all of the POIs in the given time frame for the dense environment, shown in Fig~\ref{fig:numLocal}. The green lines and blue lines do not reach 100\% by the end of the simulation. From the plots in Fig.~\ref{fig:findTimes}, the performance of the ID $A^\ast$ method and DTSPN algorithm is similar for the sparse environment. As the environment gets more complex and the error probability increases, the ID $A^\ast$ method outperforms the DTSPN, shown by lower median times to localize the POI.

\begin{figure}[h]
    \centering
    \includegraphics[width=3.25in]{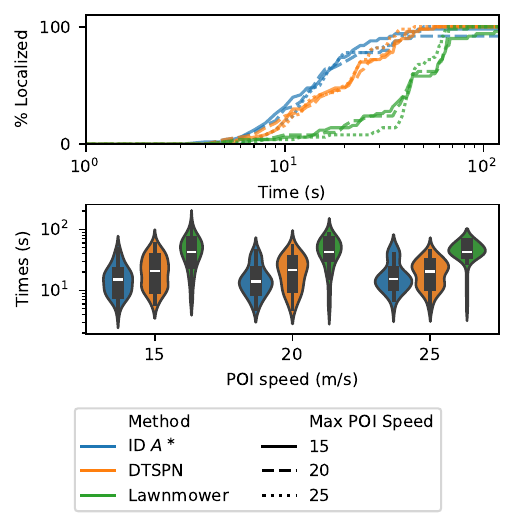}
    \caption{A plot of the time it took for each planning method to localize the POI for different POI maximum speeds.}
    \label{fig:speedTimes}
\end{figure}

Figure~\ref{fig:speedTimes} shows the percentage of POIs localized over the logarithm of time for different methods and POI maximum speeds. A smaller Monte Carlo simulation was used to test the effect of different POI speeds. The parameters were:  maximum turn-rate $u_{\psi, {\rm max}}=\pi / 4$~rad/s, probability of detection $p_d=0.8$, exploration factor $\beta = 0.8$, reward discount factor $\gamma=0.1$, initial heading $q_{\psi, 0} = \pi / 4$, max sensing distance $l_{\rm max}=300$~m, min speed $u_{v, {\rm min}}=36$~m/s, max speed $u_{v, {\rm max}}=44$~m/s, and starts in the voxel containing the $xy$ pair $[-350, -350]^{\rm T}$.
The proposed method runs every timestep until the iterative deepening algorithm plans paths for all time horizons $T=\{1, 2, 3, 5, 7, 9, 13\}$, or the planner executes for $5$~s. Three different POI maximum speeds $15$~m/s, $20$~m/s, and $25$~m/s were evaluated in the medium-density environment. The transition probabilities for the POIs with $20$~m/s and $25$~m/s maximum speed are shown in Sec.~\ref{sec:appendix}. Based on the plot in Fig.~\ref{fig:speedTimes}, different POI maximum speeds appears to affect the algorithms less than the difference in error rates in Fig.~\ref{fig:numLocal}. From the plot, the iterative deepening $A^\ast$ method finds POI more quickly than the baseline methods because the percentage of POI localized lines are above the baseline methods' corresponding lines. The second best performing algorithm was the DTSPN method, and the worst, as expected, was the lawnmower  method. The second plot in Fig.~\ref{fig:speedTimes} is violin plots with box plots inside. The medians, shown as the white line, of the proposed iterative deepening $A^\ast$ method are lower than the baseline methods. The violin plots show an increase in the density, the wider colored regions, of time to localize the POI at later times. The logarithmic time scale on the $y$-axis distorts the violin plots.

The Monte-Carlo simulation was run sequentially on an AMD Threadripper 3990X running Ubuntu 20.04. The planning time for the proposed method was capped at 5~s per timestep, the lawnmower planning method took less than 1~s, and the DTSPN method took an average of %
or 21.1~hours for the three different environments. The lawnmower method is the fastest but considers the environment's geometry the least, while the DTSPN and proposed methods both consider VVs. As expected of TSP-based approaches, the DTSPN method takes much longer to compute a path since a large number of neighborhoods are needed to cover the road network. 

%% file: sections/99_Appendix.tex
\subsection{Planning Algorithms}
Algorithm~\ref{alg:iterative} is an iterative deepening planner using $A^\ast$ to plan paths for each time horizon and return the path with the longest time horizon \cite{lavalle2006ch2}. The function takes the UAV's current position $\bm{q}_k$, the UAV's estimate of the POI $p(s|\bm{\xi}_{0:k})$, the set of time horizons $T$, and an execution time limit $\lambda$. First, the algorithm creates an empty path to store the path with the longest time horizon created during the algorithm (line~\ref{alg:iterative:initpaths}). Then, for the $i$th time horizon in the set of $T$ with $N_h$ elements, the algorithm finds a path through the environment (lines~\ref{alg:iterative:loop0}--\ref{alg:iterative:loop1}). During each loop iteration, Alg.~\ref{alg:astar} finds a path through the environment, and the path $\mathcal{P}$ is updated (line~\ref{alg:iterative:astar}). The arguments passed to the $A^\ast$ algorithm are the UAV's current position $\bm{q}_k$, the current UPV /POI probability distribution $\bm{\rho}_k$, and a subset of the time horizon $T_{0:i}$ (the time horizon for $A^{\ast}$ to plan over). Then the last tuple in the path $\mathcal{P}_i$ is checked. The tuple is the set of UAV state $\bm{q}$, timestep $\bm{\tau}$, and UPV $\bm{\rho}$ (line~\ref{alg:iterative:pop}). The early stopping condition is that the final node in the path's UPV is $\bm{0}$, the zero vector, or the execution time has passed the time limit (line~\ref{alg:iterative:stop}). If the UPV is the zero vector, it indicates that the UAV has viewed the entire environment and has found the POI. 
\begin{algorithm}[h!]
  \caption{Iterative Deepening}
  \begin{algorithmic}[1]
    \STATEx \hspace{-2.1em}  {\bf function:} $\texttt{IterativeDeepening}(\bm{q}_k,\;p(s|\bm{\xi}_{0:k}),\;T,\;\lambda)$
    \STATEx \hspace{-2.1em}   {\bf input:} $\bm{q}_k$ the UAV's current position, $p(s|\bm{\xi}_{0:k})$ the UAV's estimate of the POI, $T$ the set of time horizons for the horizon planner, $\lambda$ an execution time budget.
    \STATEx \hspace{-2.1em}  {\bf output:} $\mathcal{P}$ a set of paths through the environment.
    \STATE $\mathcal{P}=\emptyset$ // Initialize the set of paths \label{alg:iterative:initpaths}
    \STATE $\bm{\rho}_k \gets [p(s_0|\bm{\xi}_{0:k}), \ldots,p(s_{\delta-1}|\bm{\xi}_{0:k})]^{\rm T}$ // Initialize  UPV
    \FOR{$i\in\rangeset{N_h}$} \label{alg:iterative:loop0}
        \STATE $\mathcal{P} \gets \texttt{AStar}(\bm{q}_k, \bm{\rho}_k,T_{0:i})$ \label{alg:iterative:astar}
        \STATE $\bm{q}, \tau, \bm{\rho} \gets \mathcal{P}_{i}$ \label{alg:iterative:pop}
        \IF{$\bm{\rho} = \bm{0}~{\rm \textbf or}~\texttt{ElapsedTime}() \geq \lambda$} \label{alg:iterative:stop}
            \STATE $\RETURN \mathcal{P}$
        \ENDIF
    \ENDFOR \label{alg:iterative:loop1}
    \STATE $\RETURN \mathcal{P}$
  \end{algorithmic}
  \label{alg:iterative}
\end{algorithm}

Algorithm~\ref{alg:astar} is an implementation of the $A^\ast$ algorithm \cite{lavalle2006ch2}. The algorithm takes the current configuration $\bm{q}_k$, the current UPV $\bm{\rho}_k$, and a set of timesteps $T = \{\tau_0, \ldots, \tau_f\}$ where $\tau_f$ is the final timestep. The algorithm returns a path from the initial node, with the current configuration and reward, to the goal node. The goal node in this implementation is an abstract node without a state that signifies the end of planning. The goal node connects to all nodes at the last timestep in the time horizon with a zero cost edge.
First, the algorithm initializes the lookup table of node parents $\chi$ and the lookup table of best cost to a node $\omega$ (lines~\ref{alg:astar:initparents}--\ref{alg:astar:initcost}).
Then, the root node $(\bm{q}_k, \bm{\rho}_k, 0)$ is added to the priority queue $\mathcal{Q}$ (line~\ref{alg:astar:pqinit}). Next is the loop that runs until the priority queue is empty---indicating that no path to the goal was found---or the goal node is removed from the priority queue (lines~\ref{alg:astar:while0}--\ref{alg:astar:while1}).
During each iteration of the while loop, the current node with the least estimated total cost to the goal is removed from the queue (line~\ref{alg:astar:pop}), the current node is checked to see if it is the goal node (line~\ref{alg:astar:goal}), and the nodes neighboring the current node are expanded and added to the priority queue (lines~\ref{alg:astar:for0}--\ref{alg:astar:for1}).
If the current node is the goal node, the current node's timestep is not in the set of timesteps $T$ and the algorithm returns the path from the start node to the goal node using the lookup table of parent nodes $\chi$ in the $\texttt{GetPath}$ function. When expanding the neighbors of the current node with Alg.~\ref{alg:neighbors}, the cost of traveling from the start node to each child node is calculated using \eqref{eq:transition} (line~\ref{alg:astar:cost}). Alg.~\ref{alg:astar} then checks to see if the estimated cost of the path $c_c + r(\bm{q}_c, \bm{\rho}_c)$---the cost to travel to the current node and the estimated cost of traveling to the goal \eqref{eq:heuristic}---is better then the previously estimated cost of the path. The $\texttt{Get}$ function determines if the child node $(\bm{q}_c, \bm{\rho}_c, \tau_c)$ is in the lookup table of costs to travel to existing nodes $\omega$ and returns $\infty$ if it is not in the lookup table. If the estimated cost of the path is better than the previously estimated cost of the path, then Alg.~\ref{alg:astar} updates the estimated cost of the path, updates the child node's parent to the current node, and inserts the child to the priority queue or updates the child's priority in the queue with the $\texttt{Upsert}$ function (lines~\ref{alg:astar:update0}--\ref{alg:astar:update1}).

\begin{algorithm}[h!]
  \caption{$A^*$}
  \begin{algorithmic}[1]
    \STATEx \hspace{-2.1em}  {\bf function:} $\texttt{AStar}(\bm{q}_k,\;\bm{\rho}_k,\;T)$
    \STATEx \hspace{-2.1em}   {\bf input:} $\bm{q}_k$ the UAV's current position, $\bm{\rho}_k$ the UAV's UPV, $T = \{\tau_0, \ldots, \tau_f\}$ the set of time horizons for the horizon planner.
    \STATEx \hspace{-2.1em}  {\bf output:} $\mathcal{P}$ a path through the environment.
    \STATE $\chi\gets\emptyset$ // Initialize lookup table of node parents \label{alg:astar:initparents}
    \STATE $\omega \gets \emptyset$ // Initialize lookup table of best cost to a node  \label{alg:astar:initcost}
    \STATE $\mathcal{Q}=\{(\bm{q}_k,\bm{\rho}_k, \tau_0, 0)\}$ // Initialize priority queue \label{alg:astar:pqinit}
    \WHILE{$\mathcal{Q}\neq \emptyset$} \label{alg:astar:while0}
        \STATE $\bm{q}_p, \bm{\rho}_p, \tau_p, c_p \gets \texttt{Pop}(\mathcal{Q})$  // Remove node with least estimated cost \label{alg:astar:pop}
        \IF{$\tau_p \notin T$} \label{alg:astar:goal} // Check if removed node is a goal node
            \STATE $\RETURN \texttt{GetPath}(\mathcal{\chi}, (\bm{q}_p, \bm{\rho}_p, \tau_p))$ \label{alg:astar:path}
        \ENDIF
        \FOR{$\bm{q}_c, \tau_c, \bm{\rho}_c \in \texttt{Neighbors}(\bm{q}_p, \bm{\rho}_p, \tau_p, T)$}  \label{alg:astar:for0}
            \STATE $c_c = c_p + c(\bm{q}_c, \bm{\rho}_c)$ // Cost to transition to child node \label{alg:astar:cost}
            \IF{$\texttt{Get}(\omega, (\bm{q}_c, \bm{\rho}_C, \tau_c)) > c_c+r(\bm{q}_c, \bm{\rho}_c)$}  \label{alg:astar:bestcost}
                \STATE $\texttt{Set}\left(\omega,(\bm{q}_c, \bm{\rho}_C, \tau_c), c_c + r(\bm{q}_c, \bm{\rho}_c)\right)$ // Update best cost to node \label{alg:astar:update0}
                \STATE $\texttt{Set}\left(\chi,(\bm{q}_c, \bm{\rho}_c, \tau_c), (\bm{q}_c, \bm{\rho}_C, \tau_c)\right)$ // Update node parents list
                \STATE $\texttt{Upsert}(\mathcal{Q}, (\bm{q}_0, \bm{\rho}_c, \tau_c, c_c), c_c+r(\bm{q}_c, \bm{\rho}_c))$ \label{alg:astar:update1} // Update priority queue
            \ENDIF
        \ENDFOR\label{alg:astar:for1}
    \ENDWHILE\label{alg:astar:while1}
    \STATE $\RETURN \emptyset$
  \end{algorithmic}
  \label{alg:astar}
\end{algorithm}

\subsection{POI Transition Probabilities}
\begin{figure}[h]
    \centering
    \includegraphics{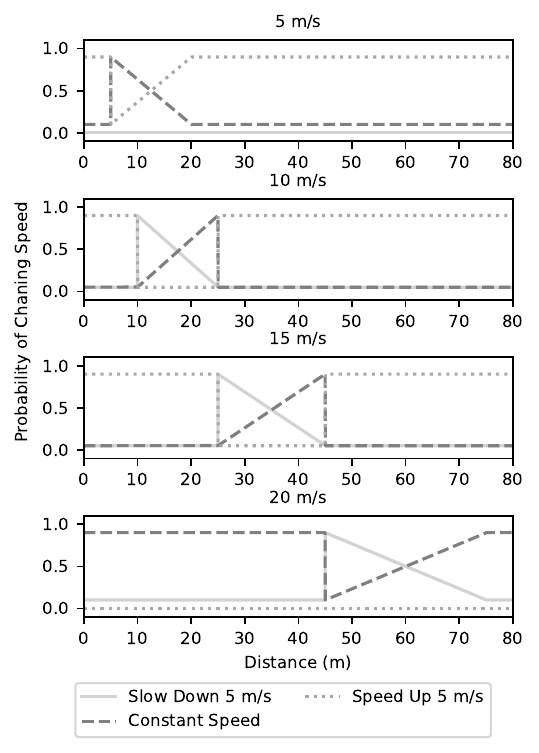}
    \caption{POI speed transition probabilities 20~m/s.}
    \label{fig:20msAccel}
\end{figure}

\begin{figure}[h]
    \centering
    \includegraphics{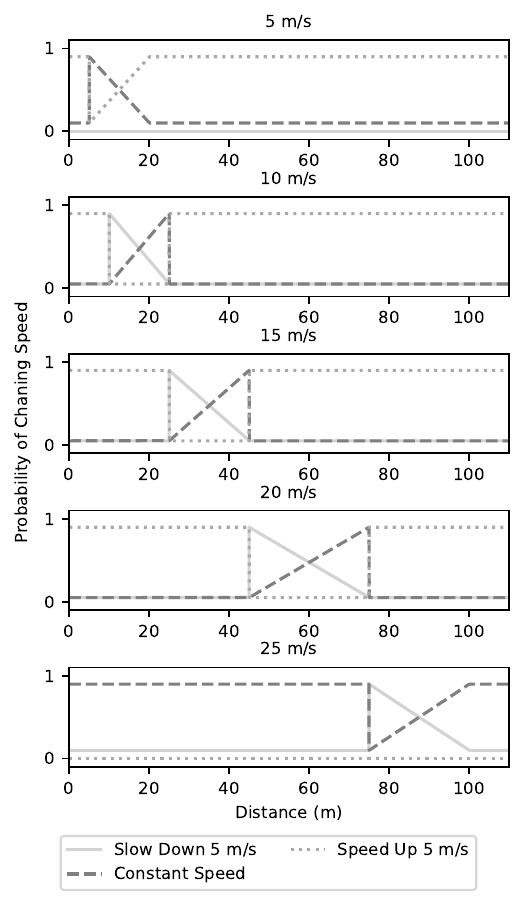}
    \caption{POI speed transition probabilities 25~m/s.}
    \label{fig:25msAccel}
\end{figure}

The transition probabilities for a POI with a maximum speed of $20$~m/s and $25$~m/s. For the Monte-Carlo simulation in Fig.~\ref{fig:speedTimes}. Table~\ref{tbl:20ms} shows the probability the POI maneuvers at various speed for the intersection in Fig.~\ref{fig:cells} when the POI has a maximum speed of 20~m/s. The probability of changing speed when approaching an intersection for a POI with a maximum speed of $20$~m/s is show in Fig.~\ref{fig:20msAccel}. The corresponding data for a POI with a maximum speed of $25$~m/s is given in Table~\ref{tbl:25ms} and Fig.~\ref{fig:25msAccel}.
\begin{table}[h]
    \centering
    \small
    \caption{The POI's probability of going in any direction at an intersection based on the current speed with a maximum speed of 20 m/s.}
    \label{tbl:20ms}
    \begin{tabular}{ccccc}
        \hline\hline
        & 5 (m/s) & 10 (m/s) & 15 (m/s) & 20 (m/s) \\
        \hline
        $\epsilon_0$ & 0.05 & 0.0 & 0.0 & 0.0\\
        $\epsilon_1$ & 0.225 & 0.15 & 0.075 & 0.025\\
        $\epsilon_2$ & 0.5 & 0.7 & 0.85 & 0.95 \\
        $\epsilon_3$ & 0.225 & 0.15 & 0.075 & 0.025 \\
        $\epsilon_4$ & 0.05 & 0.0 & 0.0 & 0.0 \\
        $\epsilon_5$ & 0.475 & 0.5 & 0.5 & 0.5 \\
        $\epsilon_6$ & 0.475 & 0.5 & 0.5 & 0.5 \\
        $\epsilon_7$ & 0.05 & 0.0 & 0.0 & 0.0 \\
        $\epsilon_8$ & 0.575 & 0.75 & 0.875 & 0.95 \\
        $\epsilon_9$ & 0.375 & 0.25 & 0.125 & 0.05 \\
        \hline
    \end{tabular}
\end{table}

\begin{table}[h]
    \centering
    \small
    \caption{The POI's probability of going in any direction at an intersection based on the current speed with a maximum speed of 25 m/s.}
    \label{tbl:25ms}
    \begin{tabular}{cccccc}
        \hline\hline
        & 5 (m/s) & 10 (m/s) & 15 (m/s) & 20 (m/s) & 25 (m/s) \\
        \hline
        $\epsilon_0$ & 0.05 & 0.0 & 0.0 & 0.0 & 0.0 \\
        $\epsilon_1$ & 0.225 & 0.15 & 0.075 & 0.025 & 0.0 \\
        $\epsilon_2$ & 0.5 & 0.7 & 0.85 & 0.95 & 1.0 \\
        $\epsilon_3$ & 0.225 & 0.15 & 0.075 & 0.025 & 0.0 \\
        $\epsilon_4$ & 0.05 & 0.0 & 0.0 & 0.0 & 0.0 \\
        $\epsilon_5$ & 0.475 & 0.5 & 0.5 & 0.5 & 0.5 \\
        $\epsilon_6$ & 0.475 & 0.5 & 0.5 & 0.5 & 0.5 \\
        $\epsilon_7$ & 0.05 & 0.0 & 0.0 & 0.0 & 0.0 \\
        $\epsilon_8$ & 0.575 & 0.75 & 0.875 & 0.95 & 1.0 \\
        $\epsilon_9$ & 0.375 & 0.25 & 0.125 & 0.05 & 0.0 \\
        \hline
    \end{tabular}
\end{table}

